# Efficient Implementation of the Plan Graph in STAN

**Derek Long** D.P.LONG@DUR.AC.UK
**Maria Fox** MARIA.FOX@DUR.AC.UK
*Department of Computer Science*
*University of Durham, UK*

## Abstract

STAN is a Graphplan-based planner, so-called because it uses a variety of STate ANalysis techniques to enhance its performance. STAN competed in the AIPS-98 planning competition where it compared well with the other competitors in terms of speed, finding solutions fastest to many of the problems posed. Although the domain analysis techniques STAN exploits are an important factor in its overall performance, we believe that the speed at which STAN solved the competition problems is largely due to the implementation of its plan graph. The implementation is based on two insights: that many of the graph construction operations can be implemented as bit-level logical operations on bit vectors, and that the graph should not be explicitly constructed beyond the fix point. This paper describes the implementation of STAN's plan graph and provides experimental results which demonstrate the circumstances under which advantages can be obtained from using this implementation.

## 1. Introduction

STAN is a domain-independent planner for STRIPS domains, based on the graph construction and search method of Graphplan (Blum & Furst, 1997). Its name is derived from the fact that it performs a number of preprocessing analyses, or STate ANalyses, on the domain before planning, using the Type Inference Module TIM described by Fox and Long (1998).

STAN competed in the AIPS-98 planning competition and achieved an excellent overall performance in both rounds. The results of the competition, which can be found at the URL given in Appendix A, show that STAN was able to solve some problems notably quickly and that it could find optimal parallel solutions to some problems which could not be solved optimally by any other planner in the competition, for example in the Gripper domain. The problems posed in the competition did not give STAN much opportunity to exploit its domain analysis techniques, so this performance is due mainly to the underlying implementation of the plan graph that STAN constructs and searches. A more detailed discussion of the competition, from the competitors' point of view, is in preparation.

The design of STAN's plan graph is based on two insights. First, we observe that action pre- and post-conditions can be represented using bit vectors. Checking for mutual exclusion between pairs of actions which directly interact can be implemented using logical operations on these bit vectors. Mutual exclusion (mutex relations) between facts can be implemented in a similar way. In order to best exploit the bit vector representation of information we construct a two-layer graph called a *spike* which avoids unnecessary copying of data and allows layer-dependent information about a node to be clearly separated from layer-*in*dependent information about that node. The spike allows us to record mutex relations using bit vectors, making mutex testing for *indirect* interaction much more efficient (we distinguish between direct and indirect interaction in Section 2.1). Second, we observe that there is no advantage in explicit construction of the graph beyond the stage at which the fix point is reached. Our plan graph maintains a *wave front* which keeps track of all of the goal sets remaining to be considered during search. Since no new facts, actions or mutex relations are added beyond the fix point these goal sets can be considered without explicit copying of the fact and action layers. The wave front mechanism allows STAN to solve very large problem instances using a fraction of the time and space consumed by Graphplan and IPP (Koehler, Nebel, & Dimopoulos, 1997). For example, using a heuristic discussed in Section 5.1, STAN can solve the 10-disc Towers of Hanoi problem (a 1023 step plan) in less than 9 minutes.

In this paper we describe the spike and wave front mechanisms and provide experimental results indicating the performance advantages obtained.

## 2. The Spike Graph Structure

Graphplan (Blum & Furst, 1997) uses constraint satisfaction techniques to search a layered graph which represents a compressed reachability analysis of a domain. The layers correspond to snapshots of possible states at instants on a time line from the initial to the goal state. Each layer in the graph comprises a set of facts that represents the union of states reachable from the preceding layer. This compression guarantees that the plan graph can be constructed in time polynomial in the number of action instances in the domain. The expansion of the graph, from which solutions can be extracted, is partially encoded in binary mutex relations computed during the construction of each layer. STAN implements an efficient representation of the graph in which a *wave front*, discussed in Section 4, further supports its compression. In Graphplan-style planners the search for a plan, from layer $k$, involves the selection and exploration of a collection of action choices to see whether a plan can be constructed, using those actions at the $k$th time step. If no plan is found the planner backtracks over the action choices. Two important landmarks arise during the construction of the plan graph. The first is the point at which the graph *opens* in the sense that the problem becomes, in principle, solvable. This is the layer at which all of the top level goals first become pairwise non-mutex and is referred to here as the *opening layer*. The second is the *fix point*, referred to as *level off* by Blum and Furst (1997), the layer after which no further changes can be made to either the action, fact or mutex information recorded in the graph at each layer.

In the original implementation of Graphplan the graph was implemented as an alternating sequence of layers of fact nodes and action nodes, with arcs connecting actions to their preconditions in the previous layer and their postconditions in the subsequent layer. The layers were constructed explicitly involving the repeated copying of large portions of the graph at each stage in maintaining the graph structure. This copying was due to two features of the graph. First, since actions with satisfied preconditions in one layer continue to have satisfied preconditions in all subsequent layers, actions that have once been added to a layer will appear in every successive action layer with the same name and the same pre- and post-conditions. Second, since facts that have once been achieved by the effects of an action will always be achieved by that action, they will continue to appear in every successive fact layer after the layer in which they first appeared. Although the layers can get deeper at every successive stage they each duplicate information present in the previous layer, so there is only a small amount of new information added at every stage. The proportion of new material, relative to copied material, decreases progressively as the graph develops.

In the original Graphplan, mutex relations were checked for by maintaining lists of facts, corresponding to the pre- and post-conditions of actions, and checking for membership of facts within these lists. Because of the need to copy information at each new layer, the pre- and post-conditions of actions were duplicated even though this information did not vary from layer to layer (it can be determined once and for all at the point of instantiation of the schema). It is possible to identify layer-independent information, with each node in the graph, which can be stored just once using a different representation of the graph structure.

The spike representation reimplements the graph as a single fact array, called the *fact spike*, and a single action array, called the *action spike*, each divided into *ranks* corresponding to the layers in the original Graphplan graph structure. The observations leading to this compressed implementation of the plan graph were made independently by Smith and Weld (1998). In Stan, a fact rank is a consecutive sequence of *fact headers* storing the layer-independent information associated with their associated facts in the corresponding fact layer. Similarly, an action rank is a consecutive sequence of *action headers* storing layer-independent information about their associated actions in the corresponding action layer. Each header is a tuple containing, amongst other things, the name of the fact or action it is associated with and a structure which stores the layer-dependent information relevant to that fact or action. In the case of fact headers this structure is called a *fact level package* and in the case of action headers it is an *action level package*. Figure 1 shows how a simple graph structure can be viewed as a spike.

In the spike the positions of all fact and action headers are fixed and can be referred to by indexing into the appropriate array. At any point, the sizes of the arrays are referred to using the constant $MaxSize$, a large number setting an upper bound on the size of the spike. All of the vectors allocated are also initialised to this size, although they are used in word-sized increments. This saves the effort of re-allocating and copying vectors as the spike increases in size towards $MaxSize$. We now define the data types so far introduced.

**Definition 1** *A* spike vector *is a bit vector of size* $MaxSize$.

**Definition 2** *A* fact header *is a tuple of six components: a* name *which is the predicate and arguments that comprise the fact itself; an* index*, i, giving the position of the fact in the fact array; a* bit mask *which is a spike vector in which the ith bit is set and all other bits are unset; a* reference *identifying its achieving no-op; a spike vector* consumers *with bits set for all the actions which use this fact as a precondition and a* fact level package *storing the layer-dependent information about that fact.*

**Definition 3** *An* action header *is a tuple of eight components: the* name *of the action; an* index*, i, giving the position of the action in the action array; a* bit mask *which is a spike vector in which the ith bit is set and all other bits are unset; a flag indicating whether the action is a no-op; three spike vectors, called* precs*,* adds *and* dels *and an* action level package *storing the layer-dependent information about that action. Each bit in precs, adds*

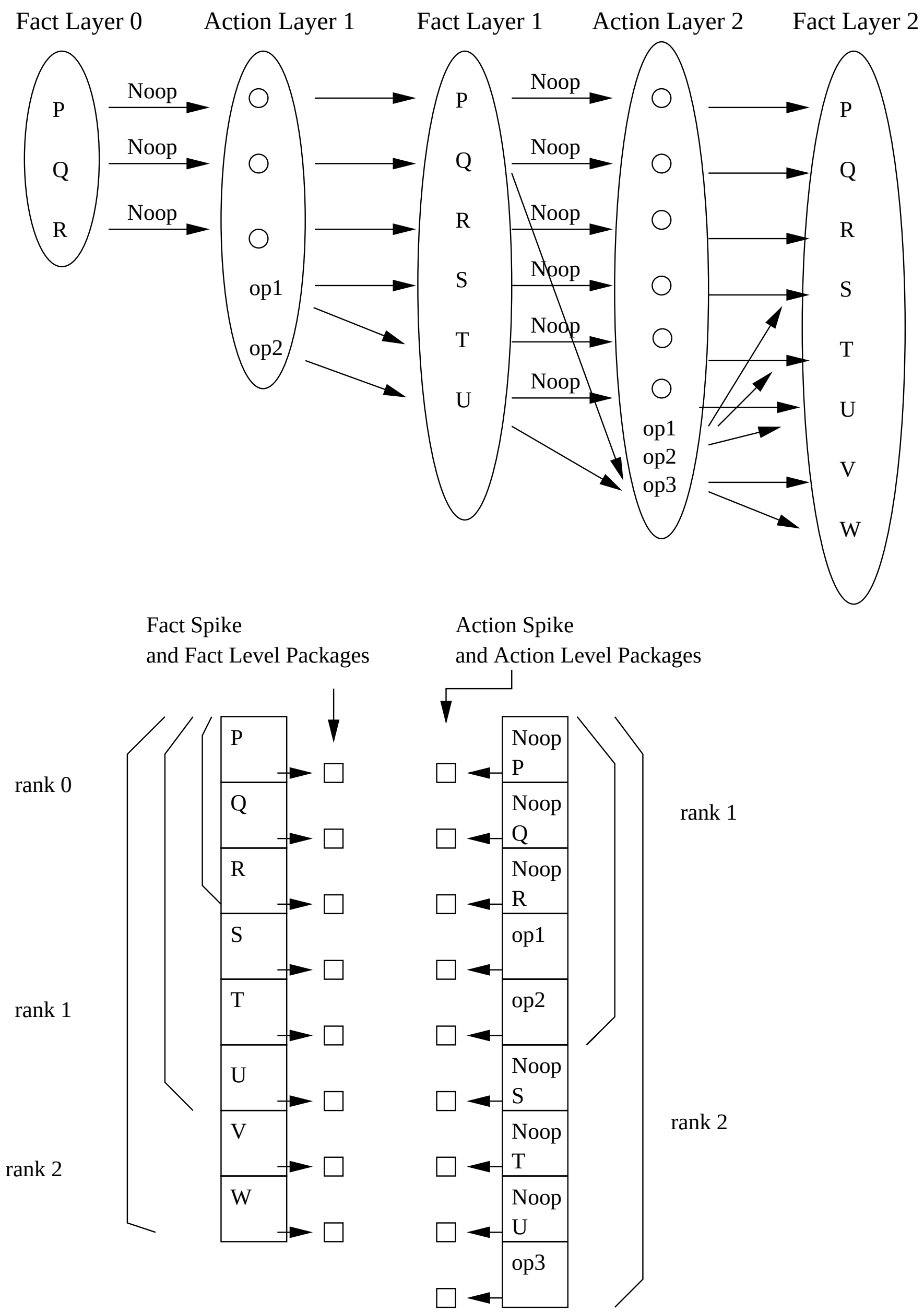

Figure 1: Representation of a plan graph as a spike. In the fact spike, ranks 0, 1 and 2 correspond to fact layers 0, 1 and 2 respectively. In the action spike, ranks 1 and 2 correspond to action layers 1 and 2 respectively.

*and dels corresponds to an index into the fact array and is set in* precs *if the fact at that index is a precondition (and unset otherwise), in* adds *if the fact at that index is an add list element (and unset otherwise) and in* dels *if the fact at that index is a delete list element of the action (and unset otherwise).*

**Definition 4** *A* fact mutex vector (FMV) *for a fact, $f$, is a spike vector in which the bits correspond to the indices into the fact array and a bit is set if the corresponding fact is mutex with $f$.*

**Definition 5** *An* action mutex vector (AMV) *for an action, $a$, is a spike vector in which the bits correspond to the indices into the action array and a bit is set if the corresponding action is mutex with $a$.*

**Definition 6** *A* fact level package *for a fact, $f$, is an array of pairs, one for each rank in the spike, each containing a* fact mutex vector *for $f$ and a vector of achievers, called the* achievement vector (AV)*, in the previous action rank.*

**Definition 7** *An* action level package *for an action, $a$, is an array of triples, one for each rank in the spike, each containing an* action mutex vector *for $a$ and a list of actions mutex with $a$* (MAs)*.*

Using these definitions we can now provide a detailed description of the spike construction process.

## 2.1 The Spike Construction Process

We will make use of these header access functions in the following discussion:

$$mvec : fact \rightarrow fact\ mutex\ vector$$
$$precs_of : action \rightarrow precs$$
$$adds_of : action \rightarrow adds$$
$$dels_of : action \rightarrow dels$$

The spike construction process takes place within a loop which stops when all goals are pairwise achievable, and thereafter alternates with search until the fix point is reached and the wave front mechanism takes over. The use of the wave front is discussed in Section 4. The key component of the process is the rank construction algorithm which builds a fact rank and an action rank by extending the previous fact and action ranks in the spike. The action rank is started by adding no-ops for each of the fact headers in the previous fact rank. As soon as these are added, the fact headers can be updated to refer, by index into the action rank, to their achieving no-ops. This information allows Stan to give preference, when searching, to plans that use the no-op to achieve a goal rather than some other achiever. In Graphplan this preference was ensured by keeping all of the no-ops at the top of the graph layers and considering the achievers in order during search.

All possible action instances are then considered. All applicable action instances are *enacted* and then removed so that they will never be reconsidered for enactment. We then identify mutex relations between the actions in the new action rank, and between facts in the new fact rank.

As in Graphplan, an action instance is applicable in a rank if all of its preconditions are present and non-mutex in the previous rank. The way in which preconditions are tested for mutual exclusion in STAN is based on our bit vector representation of fact mutex relations. We take the logical *or* of all of the fact mutex vectors of the preconditions, and logically *and* the result with the precondition vector of the action. If the result is non-zero then there are mutex preconditions and the action is not applicable. This test corresponds to checking whether the action being considered is mutex with itself - a condition we define as being *self-mutex*.

**Definition 8** *An action a, with preconditions* $a_{p1}..a_{pn}$*, is* self-mutex *if:*

$$(mvec(a_{p1}) \vee mvec(a_{p2}) \vee ... \vee mvec(a_{pn})) \wedge precs_of(a)$$

*is non-zero.*

An applicable action is enacted by adding an action header to the new rank and setting its name to the name of the action and its bit mask to record its position in the spike. In Figures 2 and 3 no-ops are given the names of the facts they achieve and are identified as no-ops by the flag components of their headers. We allocate space for the action level package and create and set its *pres*, *adds* and *dels* vectors. We then add any new facts on the add list of the action to the corresponding new fact rank. The addition of new facts requires new fact headers to be initialised.

We then identify mutex actions and mutex facts in the new ranks. Mutex actions are identified in two phases. Actions which were non-mutex in the previous rank remain non-mutex and are not considered at this stage. First, existing action mutex relations are checked to see whether they hold in the new rank. Second, new action mutex relations must be deduced from the addition of new actions in the construction of this rank. We first consider the existing action mutex relations.

Two actions are mutex, as in Graphplan, if they have conflicting add and delete lists, conflicting precondition and delete lists or mutex preconditions. In the first two cases the actions are directly, or *permanently*, mutex and never need to be re-tested although their mutex relationship must be recorded at each rank. In the third case the actions are indirectly, or *temporarily*, mutex and must be retried at subsequent ranks. We keep track of which actions to retry in order to avoid unnecessary retesting. We confirm that two actions, $a$ and $b$, which were temporarily mutex in the previous rank are still temporarily mutex using the following logical operations on the fact mutex vectors for the action preconditions. We first logically *or* together the mutex vectors for $a$'s preconditions then *and* the result with the precondition vector for $b$. If the result is non-zero then $a$ and $b$ are mutex. This procedure, which is expressed concisely in Definition 9, is identical to that for checking whether an action is self-mutex except that, in this case, the result of *or*ing the fact mutex vectors of the preconditions of one action is *and*ed with the precondition vector of the other action. Since mutex relations are symmetric it is irrelevant which action plays which role in the test.

**Definition 9** *Two actions a (with n preconditions* $a_{p1}..a_{pn}$*) and b are* temporarily mutex *if*

$$(mvec(a_{p1}) \vee mvec(a_{p2}) \vee ... \vee mvec(a_{pn})) \wedge precs_of(b)$$

*is non-zero.*

We now consider what new mutex relations can be inferred from the introduction of the new actions. It is necessary to check all new actions against all actions in the spike. This check is done in only one direction - low-indexed actions against high-indexed actions - so that the test is done only once for each pair. We check for both permanent and temporary mutex relations. The permanent mutex test is done first, because if two actions are permanently mutex it is of no interest to find that they are also temporarily mutex. Definition 10 provides the logical operation used to confirm that two actions are permanently mutex. Temporary mutex relations are checked for using the logical operation defined in Definition 9.

**Definition 10** *Two actions a and b are* permanently mutex *if the result of*

$$((precs_of(a) \vee adds_of(a)) \wedge dels_of(b)) \vee \\ ((precs_of(b) \vee adds_of(b)) \wedge dels_of(a))$$

*is non-zero.*

We add these mutex relations by setting the appropriate bits in the mutex vectors of each of the new actions. This is done by *or*ing the mutex vector of the first action with the bit mask of the second action, and vice versa. A list of mutex actions is also maintained for use during search of the spike.

A refinement of the action mutex checking done by Stan is the use of a record of actions whose preconditions have lost mutex relations since the last layer of the graph. This record enables Stan to avoid retesting temporary mutex relations between actions when the mutex relations between their preconditions cannot have changed. We use a bit vector called *changedActs* to record this information. Each fact which loses mutex relations between layers adds its consumers to *changedActs*. The impact of this refinement on efficiency is discussed in Section 3.

This concludes the construction of the new action rank. The new fact rank has already been partially constructed by the addition to the spike of fact headers for any add list elements, of the new actions, that were not already present. Now it is necessary to determine mutex relations between all pairs of facts in the spike. To do this we must first complete the achievement vectors for all of the fact headers in the new rank. Any non-mutex pairs remain non-mutex, as with actions, so effort is focussed on deciding whether previously mutex facts are still mutex following the addition of the new actions, and whether new facts induce new mutex relations. Two facts are mutex if the only way of achieving both of them involves the use of mutex actions. We therefore consider every new fact with every other fact, in only one direction. The pair $f$, $g$ is mutex in the new rank if every possible achiever of $f$ is mutex with every possible achiever for $g$. The test for this exclusion is done using $g$'s *achievement vector* and the result of logically *and*ing the action mutex vectors for all possible achievers of $f$. the following definition gives the details:

**Definition 11** *Two facts, f and g, are* mutex *if:*

$$vec_g \wedge all_mutex_f = vec_g$$

*where $vec_g$ is g's achievement vector and $all_mutex_f$ is the consequence of* and*ing all of the action mutex vectors of all of f's possible achievers.*

It does not matter in which order $f$ and $g$ are treated. The computation of the above condition corresponds to testing the truth of

$$\forall a \cdot \forall b \cdot (achiever(a, f) \wedge achiever(b, g) \rightarrow mutex(a, b))$$

Since mutex relations are symmetric and the quantifiers can be freely reordered the expression equally corresponds to

$$vec_f \wedge all_mutex_g = vec_f$$

If $f$ and $g$ are found to be mutex then we set the fact mutex vector of $f$ by *or*ing it with $g$'s bit mask and the fact mutex vector of $g$ is set conversely. This concludes the rank construction process and one iteration of the spike construction process.

### 2.2 Subset Memoization in Stan

Most of the search machinery used in Stan is essentially identical to that of Graphplan. That is, a goal set is considered by identifying appropriate achieving actions in the previous layer and propagating their preconditions back through the graph. The use of the spike and bit vector representations does not impact on the search algorithm. We experimented with using bit vector representations of bad goal sets in the memoization process, in order to exploit logical bit operations to test for subset relations between sets of goals, but this proved too expensive and we now rely upon a trie data structure. This benefits marginally from the spike because goal sets do not need to be sorted for subset testing. The order in which the goals are generated in the spike can be taken as the canonical ordering since goal sets are formed by a simple sweep through the spike at each successive layer. Stan implements an improvement on the goal set memoization of Graphplan. In the original Graphplan, when a goal set could not be achieved at a particular layer the entire set was memoized as a bad set for that layer. In Stan version 2, only the subset of goals that have been satisfied at the point of failure, within a layer, are actually memoized. More goal sets are likely to contain the smaller memoized subset than would be likely to contain the complete original failing goal set. This therefore allows us to prune search branches earlier.

This method is a weak version of Kambhampati's (1998, 1999) EBL (Explanation-Based Learning) modifications. EBL allows the identification of the subset of a goal set that is really responsible for its failure to yield a plan. Memoization of smaller sets increases the efficiency of the planner by reducing the overhead necessary in identifying failing goal sets. DDB (Dependency-Directed Backtracking) improves the search performance by ensuring that backtracking returns to the point at which the last choice responsible for failure was made. These modifications result in smaller sets being memoized and a more efficient search behaviour which, in combination with the trie, ensure that a higher proportion of failing search paths are terminated early.

We have experimented with an implementation of the full EBL/DDB modifications proposed by Kambhampati, but there is an interaction between the EBL/DDB machinery and the wave front of Stan which we are currently attempting to resolve. Our experiments so far indicate that both the wave front and EBL/DDB have significant beneficial impact on search, but not consistently across the same problems. We believe that we can enhance the advantages of the wave front by full integration with EBL/DDB, but this remains to be demonstrated.

### 2.3 A Worked Example

We now demonstrate the spike construction process in action on a simple blocks world example in which there are two blocks and two table positions. In the initial state, both blocks are on the table, one in each of the two positions. Consequently there are no clear table positions. The initial spike consists of a fact rank containing fact headers for the four facts that describe the initial state. There is a single operator schema, $puton(Block, To, From)$, as follows:

| **puton(X,Y,Z)** | |
|---|---|
| Pre: | on(X,Z), clear(X), clear(Y) |
| Add: | on(X,Y), clear(Z) |
| Del: | on(X,Z), clear(Y) |

The action rank is initially empty. On the first iteration of the loop the first action rank is constructed by creating no-ops for every fact in the zeroth fact rank. Two further actions are applicable and are enacted, and the facts on their addlists are used to create a new fact rank. This results in the partially developed spike shown in Figure 2.

It can be observed from Figure 2 that, following enactment, the fact headers associated with the newly added facts are incomplete, and although the new fact level and action level packages have been allocated they do not yet contain any values. The new fact headers are missing references to the no-ops that will be used to achieve them in the next action rank. The new fact level packages are blank because their corresponding fact headers will have no level information for rank 0.

After identification of mutex actions and mutex facts, the picture is as shown in Figure 3. In the action level packages, the lists of mutex actions are given as lists of indices for the sake of clarity. In fact they are lists of pointers to actions, in order to avoid the indirection involved in the use of indices. None of the action pairs are temporarily mutex at rank 1 because all of the fact mutex vectors from rank 0 are zero-valued.

## 3. Empirical Results

In this section we present results demonstrating the efficiency of the spike and vector representation of the plan graph used by Stan. We consider graph construction only in this section – the efficiency of search in Stan will be demonstrated in Section 4. We show the efficiency of graph construction in Stan by showing relative performance figures for Stan and the competition version of Ipp in several of the competition domains and two further standard bench mark domains. These are the Graphplan version of the Travelling Salesman domain (Blum & Furst, 1997), which uses a complete graph and is referred to here as the Complete-Graph Travelling Salesman domain, and the Ferry domain available in the PDDL release.

We compare Stan with Ipp because, to the best of our knowledge, Ipp is the only other fast Graphplan-based planner currently publicly available. We use the competition version of Ipp because this is the most up to date version available from the Freiburg webpage at the time of writing. In order to focus on the graph construction phase, and eliminate the search phase from both planners, we have constructed versions of Stan and Ipp which terminate once the graph has opened. We have removed from Stan all of the unnecessary pre-processing, domain analysis and additional features that contribute to later search efficiency. However, since Ipp is designed to build one more layer before opening than is strictly necessary, to include a dummy goal corresponding to the achievement of the conjunction of the top level goal set, we make Stan build one extra layer too so that the two systems are comparable. We have removed all of the meta-strategy control from Ipp, forcing Ipp directly into its graph construction. It is possible that a more streamlined graph constructor could be built from Ipp by elimination of further processing, but we observed, during experimentation with Ipp, that pre-processing accounts for insignificant proportions of the timings reported below. We are therefore confident that any further streamlining would have minimal effects on our results. In order to compare Stan and Ipp accurately it was necessary to modify the timing mechanisms to ensure that precisely the same elements are timed. A Unix/Linux *diff* file is available at the Stan website, and in Online Appendix 1, for anyone interested in reconstructing the Ipp graph construction system we have used. The domains and problems used, and our graph construction version of Stan, can also be found at these locations.

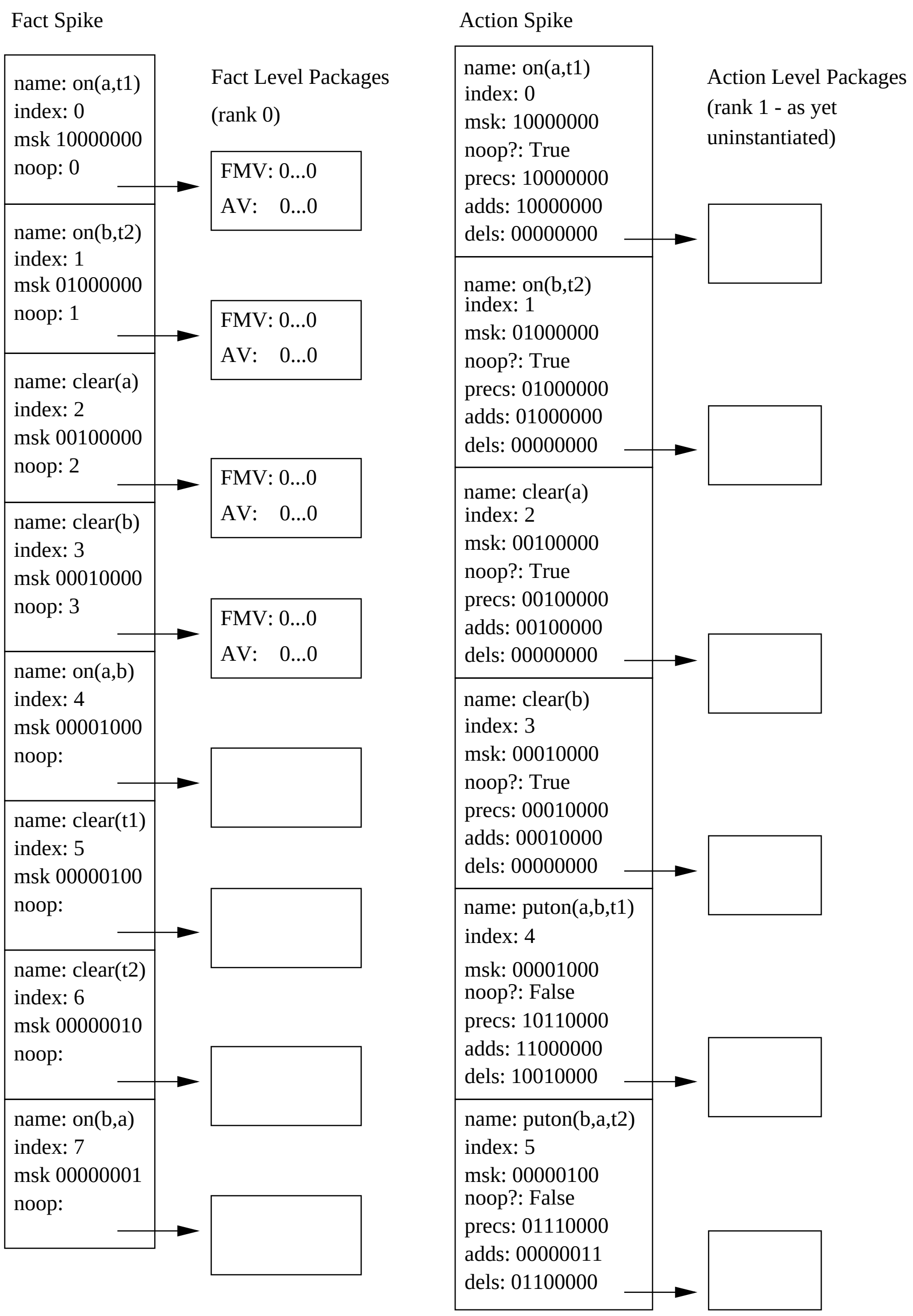

Figure 2: The spike after enactment of the rank 1 actions.

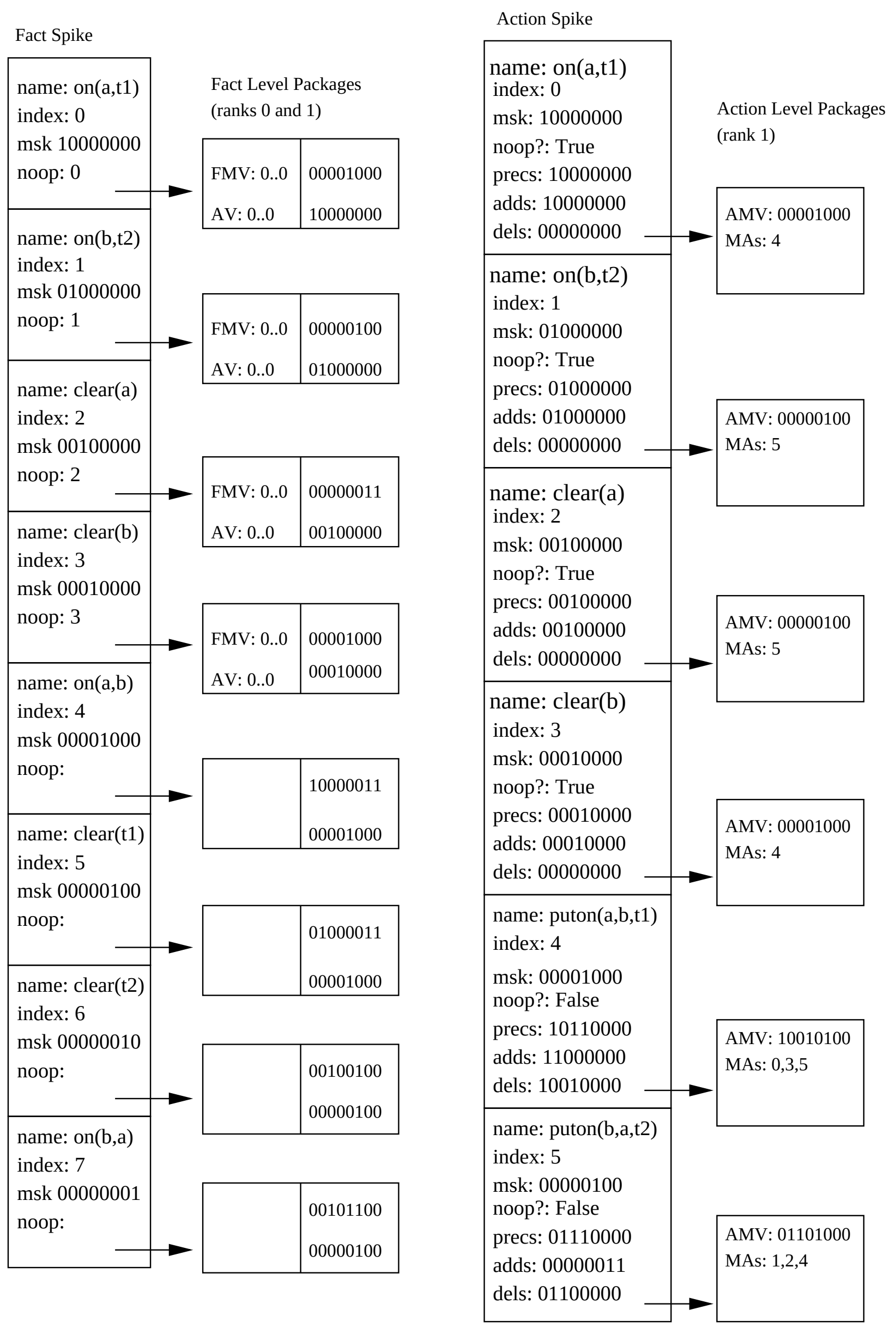

Figure 3: The spike at the end of the rank 1 construction phase.

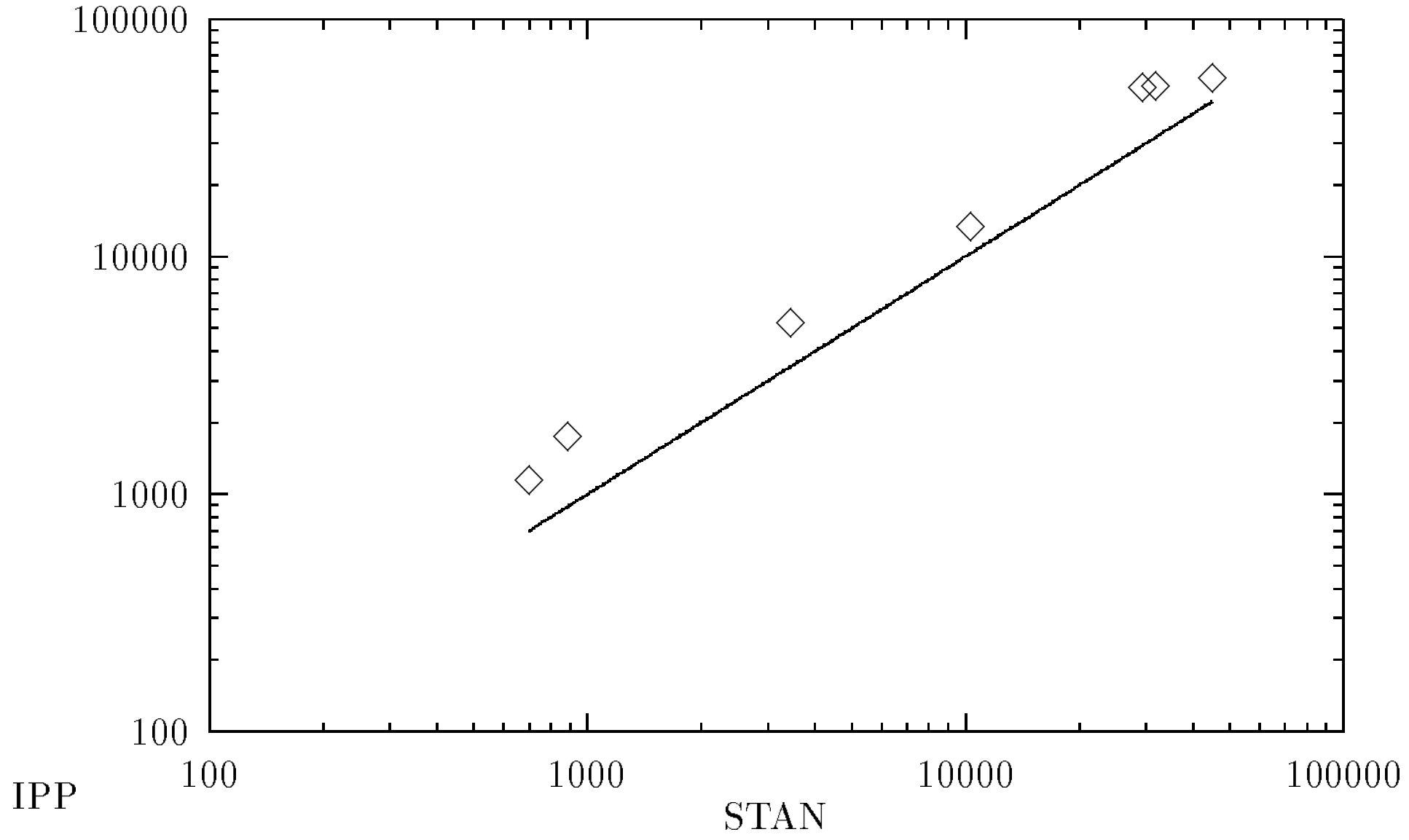

Figure 4: Graph construction in the logistics domain: Stan shows a constant factor improvement over the performance of Ipp.

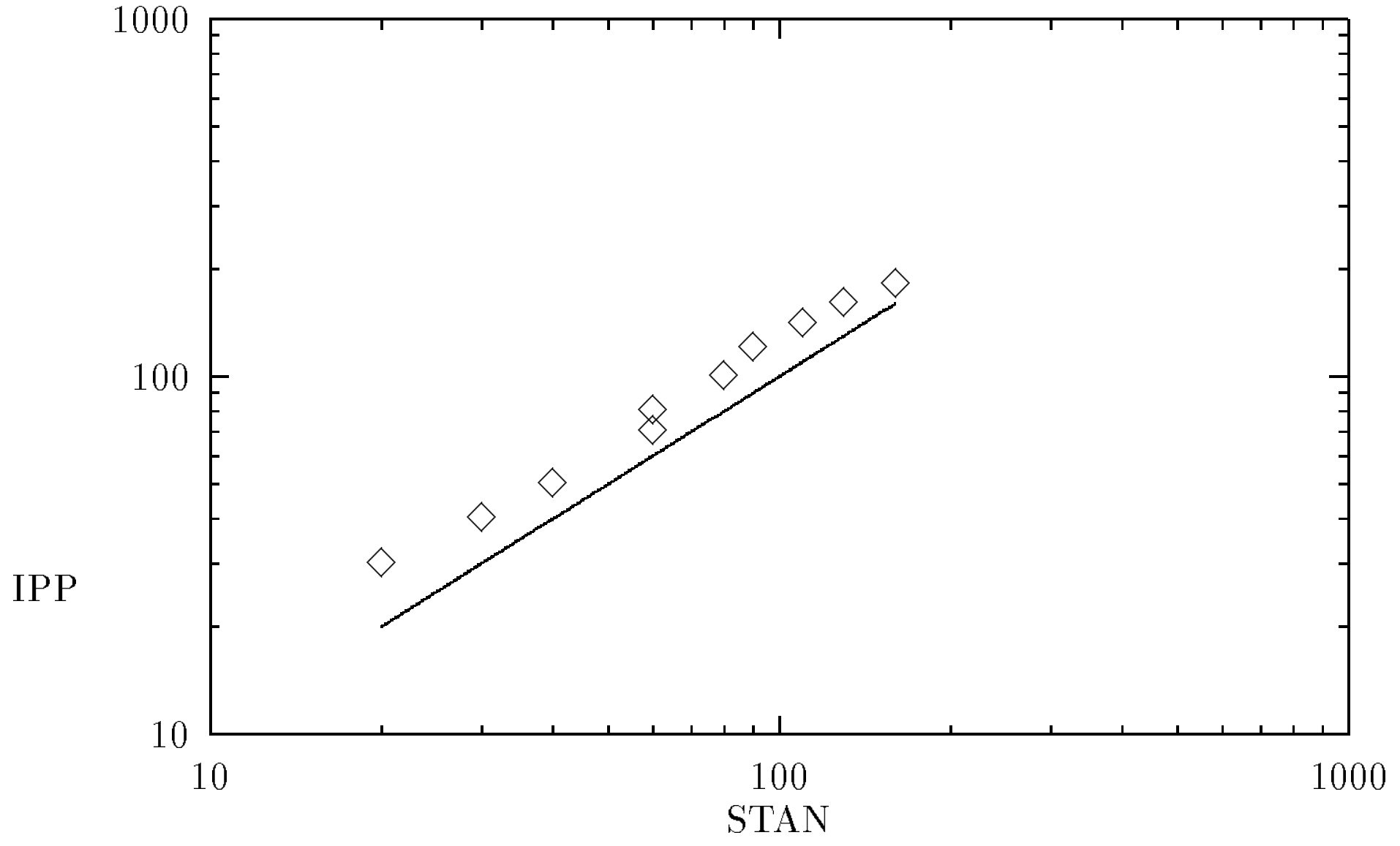

Figure 5: Graph construction in the Gripper domain.

All experiments reported in this paper were carried out on a P300 Linux PC, with 128Mb of RAM and 128Mb swap space. All of the timings in the data sets reported are in milliseconds.

All the graphs are log-log scaled. This was necessary to combat the long scales caused by very large timings associated with a few instances in each domain. The graphs show Ipp's construction performance compared with Stan's construction performance measured on the same problems in each of six domains. The straight line shows where equal performance would be. Points above the line indicate superior performance by Stan and points below the line indicate superior performance by Ipp. In all of the first five data sets, Stan clearly out-performs Ipp. In the last data set (Figure 9), Ipp convincingly out-performs Stan and we now consider a more detailed analysis of the characteristics of the domains and instances which explain these data sets.

The first four data sets reveal a very similar performance. The points are broadly parallel to the equal performance line, indicating that Stan performs at a constant multiple of the performance of Ipp. Despite the trend that these data sets reveal, occasional data points deviate significantly from this behaviour. This reflects the fact that different structures of particular problems exercise different components of the graph construction system. Components include instantiation of operators, application of individual operator instances and the corresponding extension of fact layers and checking and re-checking mutex relations between facts and between actions. We observed that in some problem instances, 50 per cent or more of the construction time was spent in action mutex checking, whilst in others instantiation dominated. The density of permanent mutex relations between actions, and the degree of persistence of temporary mutex relations between actions, are both very significant in determining efficiency of performance. For example in problem 8 in the Mystery domain, where 21 layers are constructed before the graph opens, only 9 per cent of the action pairs were discarded as permanently mutex and, of the temporary mutex pairs, the average number of re-tests across the entire graph construction was over 7. The use of the *changedActs* mechanism described in Section 2.1, to avoid retesting actions when their precondition mutex relations had not changed from the previous layer, gave us a 50 per cent improvement in performance and accounts for a more than 40 second advantage over Ipp in the construction phase of this problem.

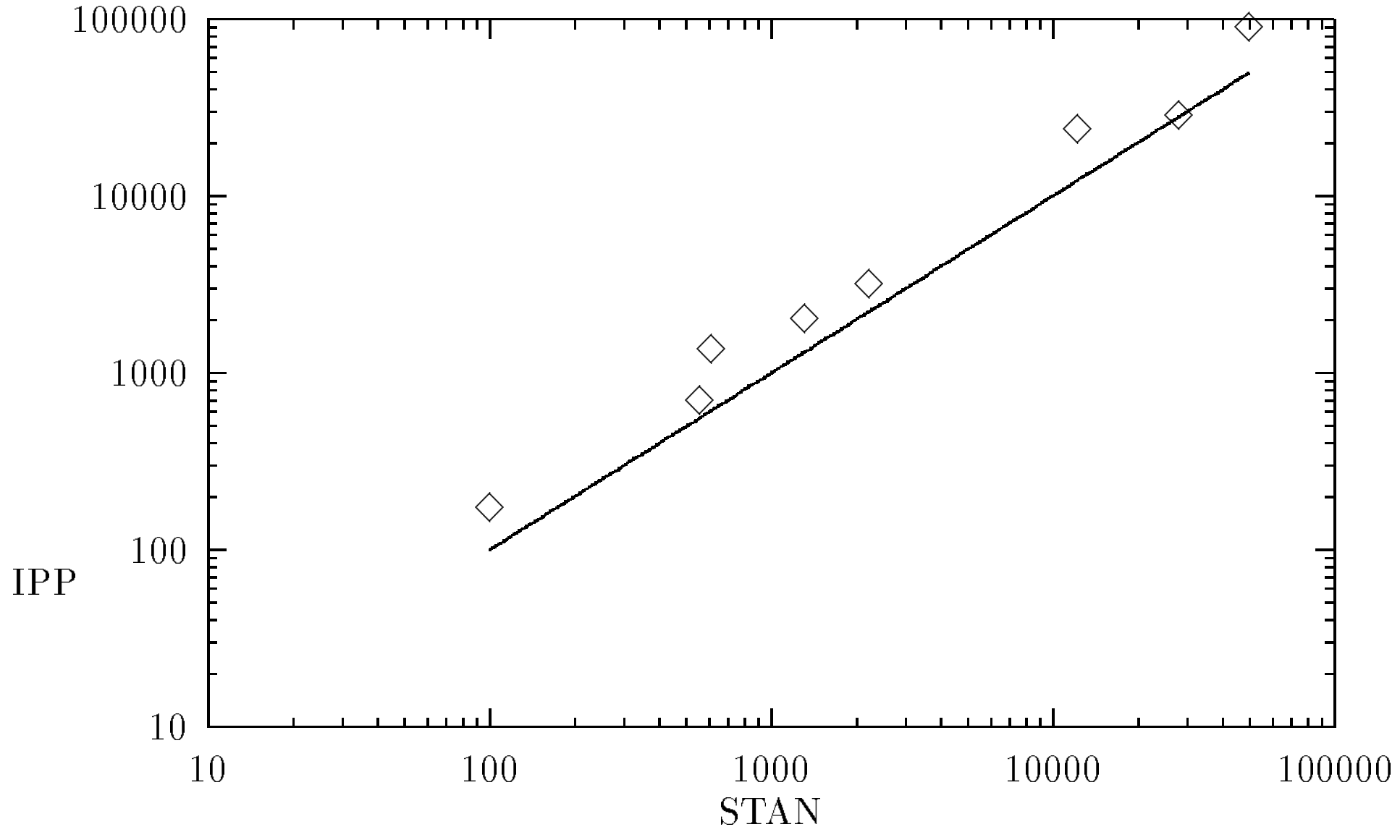

Figure 6: Graph construction in the Mystery domain. Stan's performance in this domain is consistently better than that of Ipp, but shows more marked variation revealing that the benefits of the spike are problem-dependent.

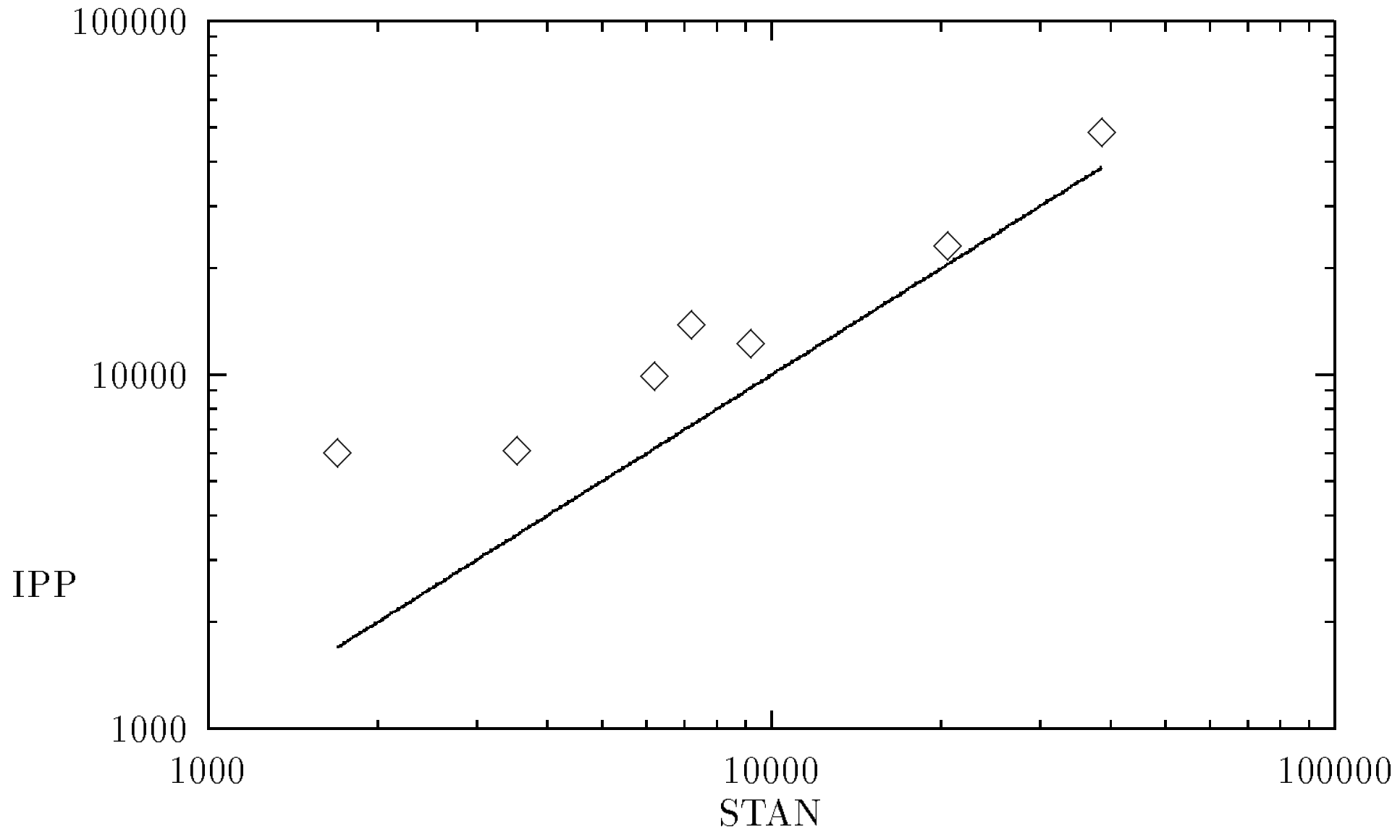

Figure 7: Graph construction in the Mprime domain.

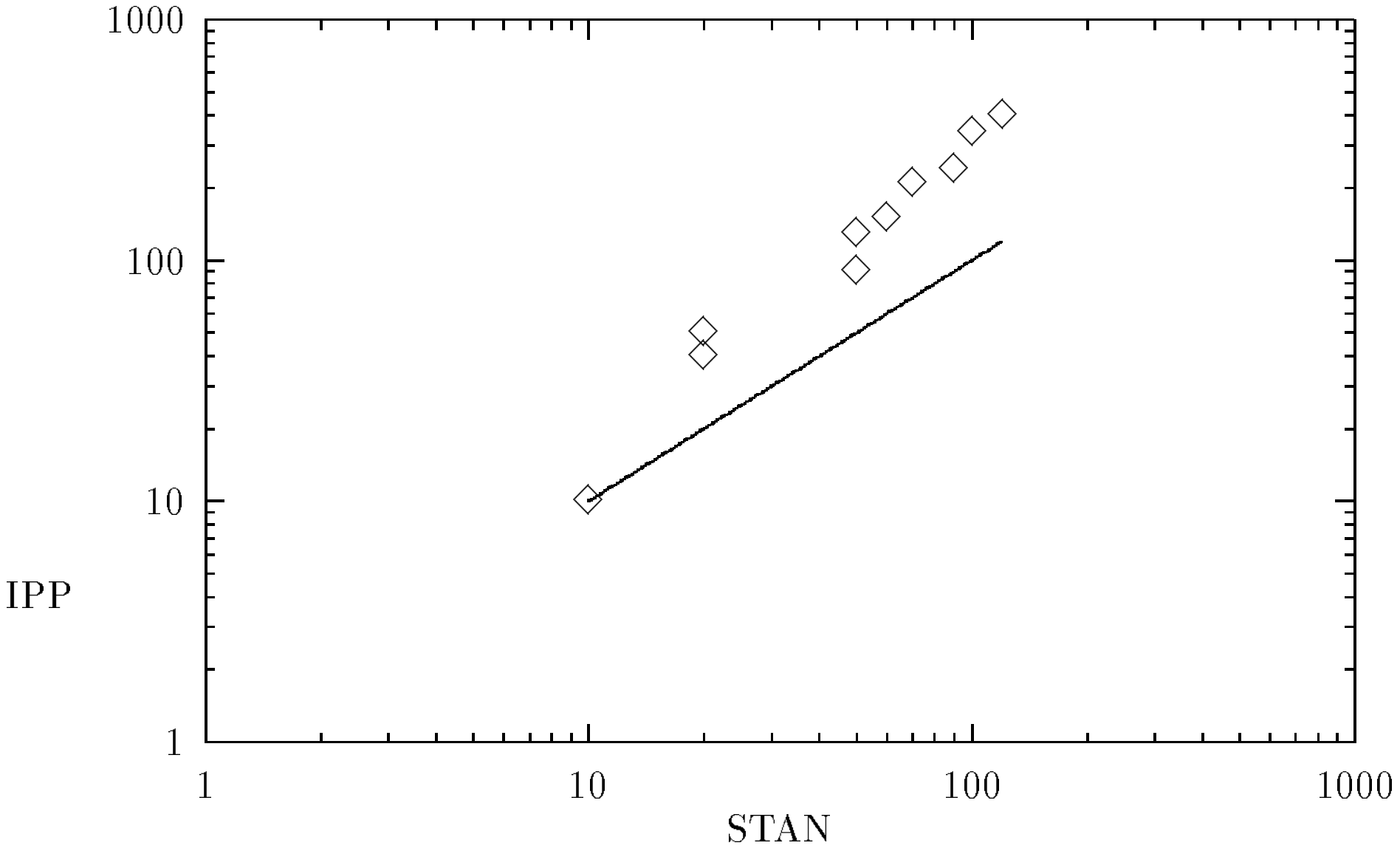

Figure 8: Graph construction in the Ferry domain. Stan shows polynomially better graph construction performance than Ipp.

In other problems a much higher percentage of action pairs are permanently mutex, allowing early elimination of many action pairs from further retesting. Where mutex relations are not highly persistent a similar elimination rate is possible. This allows much faster construction for Stan. Ipp does not benefit in the same way, because it does not distinguish between temporary and permanent mutex and does not try to identify which pairs of actions should be retested.

In the Ferry domain, Figure 8, 7 layers are constructed to open the graph regardless of instance size. Analysis reveals that approximately 25 per cent of action pairs are permanently mutex and the average persistence of temporary mutex relations is slightly more than 2 layers. Since Ipp does not intelligently eliminate actions from retesting, the implication of this is that Ipp unnecessarily re-checks mutex relations for a polynomially increasing number of pairs of actions. This explains the polynomial advantage obtained by Stan in this domain.

The last data set shows a rather different pattern of performance from that of the others. The Complete-Graph Travelling Salesman domain used to produce the data set for Figure 9 is a simplified version, in which the graph is fully connected, of the well known

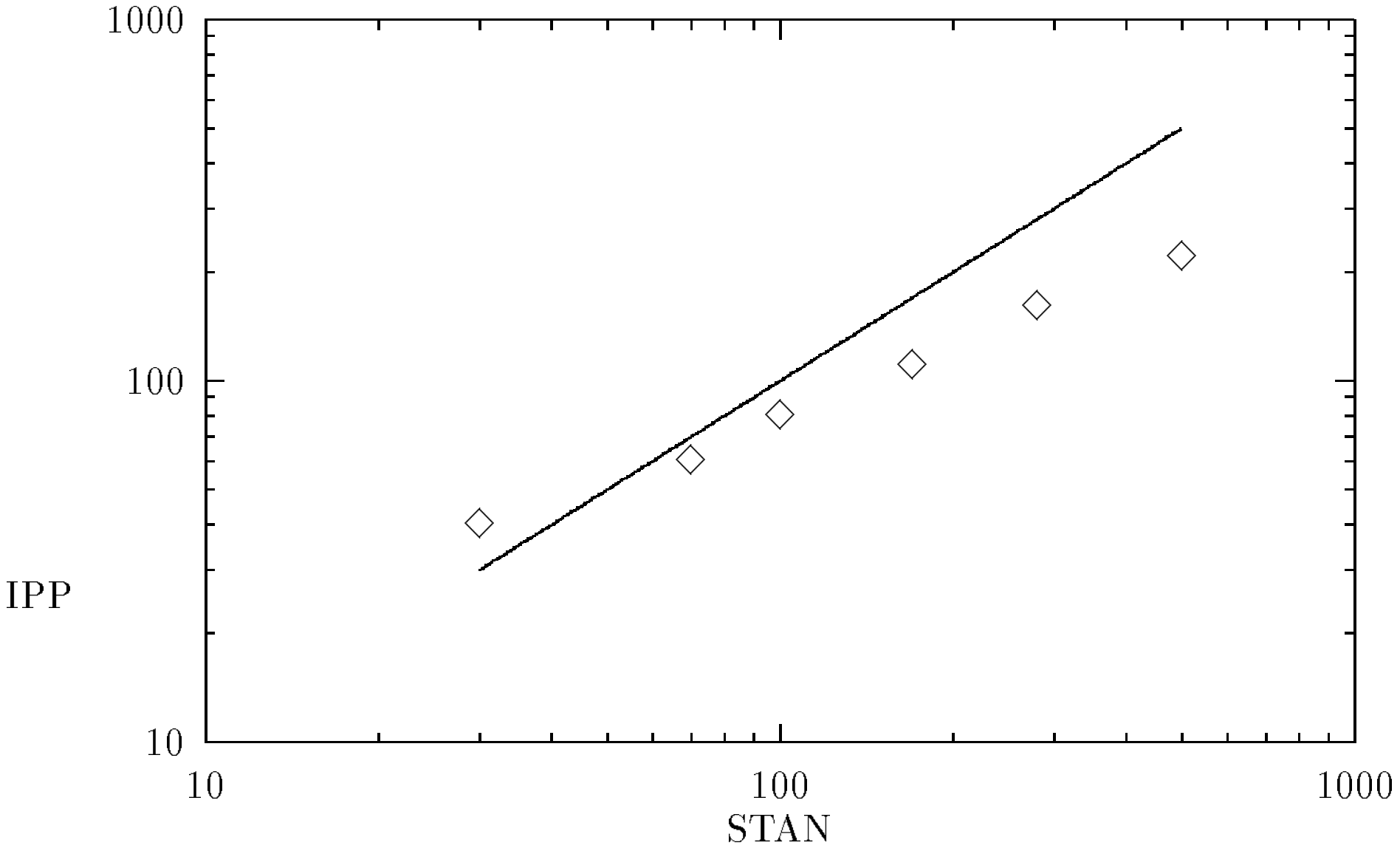

Figure 9: Graph construction in the Complete-Graph Travelling Salesman domain. Stan displays a polynomially deteriorating graph construction performance. This is further discussed in the text.

NP-hard TSP. It is, in principle, efficiently solvable. In Figure 9 Ipp's performance appears to be polynomially better than that of Stan. Analysis of the graph structure built for different instances reveals that, on all instance sizes, the graph opens at layer 3. In these graphs an interesting pattern can be observed in the mutex relations between actions: the vast majority of action pairs are mutex after their first application at layer 2 (because the salesman can only ever be in one place). These mutex relations are considered, by both Stan and Ipp, to be temporary although they in fact persist. The consequence is that both Stan and Ipp retest all pairs at the next layer. Stan obtains no advantage from the use of *changedActs* or the distinction between temporary and permanent mutex relations in this domain. The number of mutex pairs to be checked increases quadratically with increase in instance size, which is in line with Stan's performance. Ipp clearly pays much less for this retesting, despite the fact that it does the same amount of work. This fact, together with profiling of both systems, leads us to believe that the disadvantage suffered by Stan is due to the overhead in supporting object member applications in its C++ implementation. It is worth pointing out that in the Complete-Graph Travelling Salesman domain, as well as in Gripper and Ferry, the construction time for both planners is under 1 second for all instances tested so the discrepancies in performance in these three domains are insignificant compared with the discrepancies measured in seconds (for large instances) in the other domains.

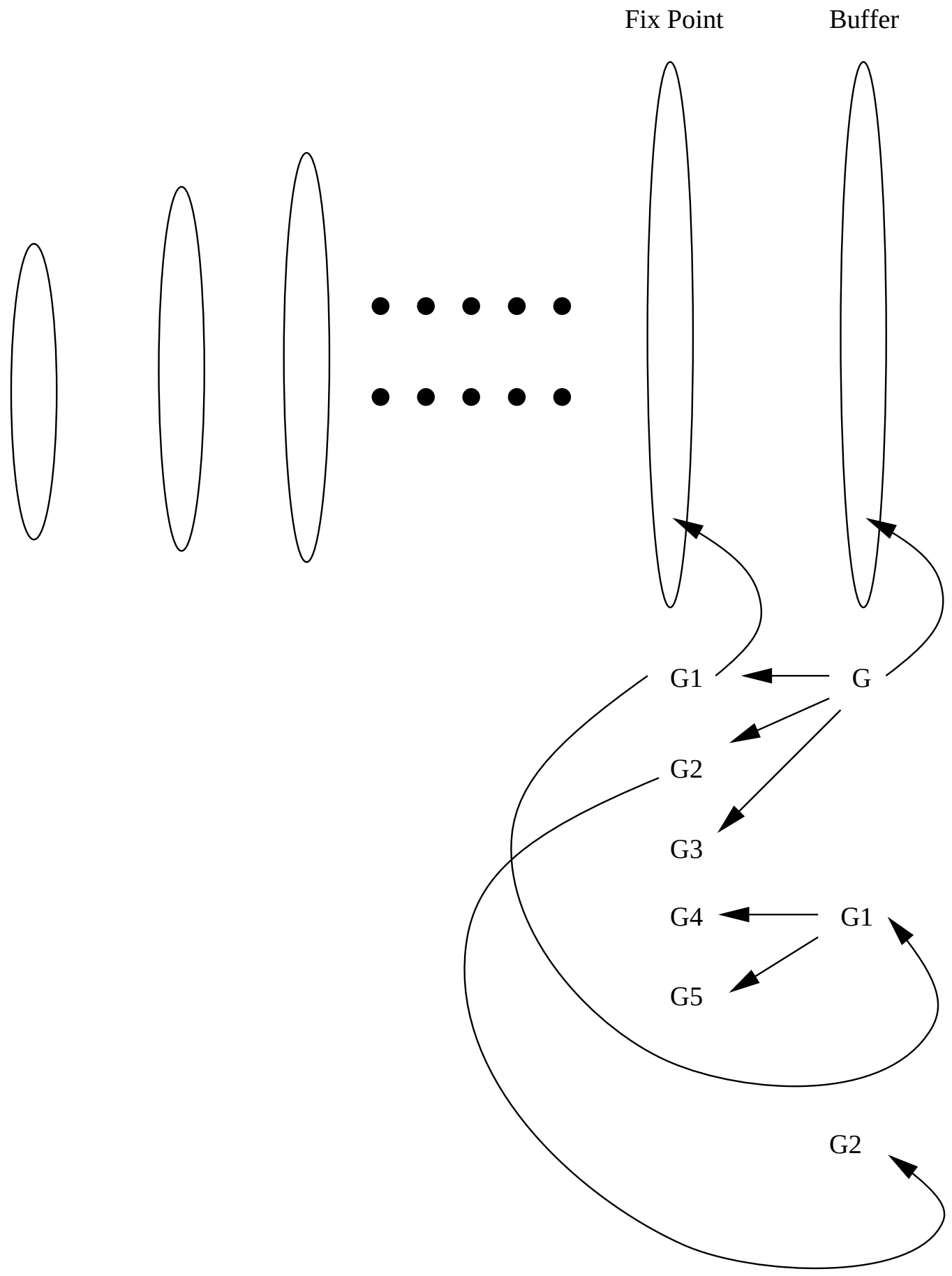

Figure 10: The wave front in Stan.

## 4. The Wave Front

When a layer is reached in which all of the top level goals are pairwise non-mutex Graphplan-based planners begin searching for a plan. If no plan can be found, new layers are constructed alternately with search until the fix point of the graph is reached. In Graphplan and Ipp the graph continues to be explicitly constructed beyond the fix point, even though the layers which can be built beyond this point are sterile (contain no new facts, actions or mutex relations). Their construction is necessary to allow the conditions for achievement of goal sets to be established, between the fix point and the current layer. However, this constitutes significant computational effort in copying existing structures and in unnecessary searching of these duplicate structures. Instead of building these sterile layers explicitly, Stan maintains a single layer, called the *buffer*, beyond the fix point together with a queue of goal sets remaining to be considered. Each time a goal set is removed from this queue, to be considered in the buffer, those goal sets it generates in the fix point layer, which have not been previously marked as unsolvable, are added to the queue. The goal sets in this queue are considered in order, always for achievement in the buffer layer. Thus, rather than constructing a new layer each time the top level goal set proves unsolvable, and then reconsidering all of the same achievers in the new layer, the goal sets in the queue are simply considered in the buffer layer. We call this mechanism a *wave front* because it pushes goal sets forward from the fix point layer into the buffer, and then recedes to consider another goal set from the fix point layer. The goal sets generated at the fix point, which join the queue for propagation, are referred to as *candidate* goal sets. The wave front is depicted in Figure 10. The underlying implementation of the plan graph remains based on the spike, but the figure depicts the graph in the traditional way for simplicity.

In the picture, G represents the top level goal set and when it is used to initiate a plan search from the buffer layer it generates the sequence of goal sets G1, G2 and G3 at the fix point layer. Assuming that these all fail, the first set in this queue, G1, is propagated forward to the buffer leading to the generation of goal sets G4 and G5 in the fix point layer. These are added to the end of the queue and G2 will be the next goal set selected from the queue to propagate forward.

In order to demonstrate that the wave front machinery maintains an appropriate behaviour there are three questions to be considered.

1. Is every goal set that would have been considered in the buffer layer, had the graph been constructed explicitly, still considered using the wave front? This question concerns completeness of the search process.
2. Does every plan generated to achieve a goal set that is considered in the buffer layer correspond to a plan that would have been generated had the graph been explicitly constructed? This question concerns soundness.
3. The final question concerns whether the termination properties of Graphplan are maintained.

**Definition 12** *A $k$-level* goal tree *for goal set $G$ at layer $n$ in a plan graph, $GT_{k,G,n}$, is a general tree of depth $k$ in which the nodes are goal sets and the parent-child relationship is defined as follows. If the goal set $x$ is in the tree at level $i$ then the goal set $y$ is a child of $x$ if $y$ is a minimal goal set containing no mutex goal pairs such that achievement of $y$ at layer $n-i-1$ in the plan graph enables the achievement of $x$ at layer $n-i$ in that graph. We take the root to be at level $0$ of the tree and the leaves to be at level $k-1$.*

**Lemma 1** *If $n-k \geq FP$ then $GT_{k,G,n} = GT_{k,G,n+1}$, where $FP$ is the number of the fix point layer in the plan graph.*

**Proof** By definition of the fix point, all layers in a plan graph beyond the fix point contain an exact replica of the information contained at the fix point layer. Since, by definition of the goal tree, the parent-child relationship depends exclusively upon the relationship between two consective layers in the plan graph, and layers cannot change after the fix point, it follows that if $x$ is the parent of $y$ at some layer beyond the fix point then the parent-child relationship between $x$ and $y$ must hold at any pair of consecutive layers beyond the fix point. Further, no new parent-child relationships can arise beyond the fix point. The restriction that $n-k \geq FP$ ensures that all layers in both goal trees lie in the region beyond the fix point.

□

The completeness of Stan follows from the completeness of Graphplan provided that all of the goal sets that would appear in the layer after the fix point in the explicit graph arise as candidates to be considered in the buffer layer using the wave front. We now prove that this condition is satisfied by first proving that the leaves of goal trees generated at successive layers of a plan graph are all used to generate candidates in Stan. Since the goal sets considered by Graphplan are always subsets of the leaves of goal trees it will be shown that the completeness of Stan follows.

**Theorem 1** *Given a goal set, $G$, and a plan graph of $n$ layers, containing no plan for $G$ of length $n-1$, with fix point at layer $FP$ ($n > FP$), all of the leaves of $GT_{n-FP,G,n}$ are generated as candidates by* Stan.

**Proof** The proof is by induction on $n$, with base case $n = FP + 1$. In the base case the result follows trivially because the only leaf in $GT_{1,G,FP+1}$ is the top level goal set $G$ and this is generated as the initial candidate by Stan.

Suppose $n > FP + 1$. The inductive hypothesis states that all of the leaves of the tree $GT_{n-1-FP,G,n-1}$ are generated as candidates by Stan. Since the plan graph constructed by Stan is identical to that of Graphplan up to layer $FP + 1$, and all candidates are used to initiate search from layer $FP + 1$, the leaves of $GT_{n-FP,G,n-1}$ will also be generated as goal sets in layer $FP$ by Stan. These goal sets are then used by Stan to construct candidates. Stan will not generate multiple copies of candidates, but each new goal set will generate a new candidate.

By Lemma 1, $GT_{n-FP,G,n} = GT_{n-FP,G,n-1}$, so that the leaves of $GT_{n-FP,G,n}$ are generated as candidates by Stan.

□

The definition of goal trees captures precisely the relationship between goal sets and the search paths considered by Graphplan. However, because Graphplan memoizes failed goal sets it can prune parts of a goal tree as it regresses through the explicit plan graph during search. Whenever a goal set contains a memoized goal set search terminates along this branch and none of its children will be generated. It can now be seen that Graphplan will generate at layer $FP + 1$ a subset of the leaves of $GT_{n-FP,G,n}$, when searching from layer $n$ with goal set $G$, whereas Theorem 1 demonstrates that Stan will construct all of these leaves as candidates.

This argument might suggest that Stan engages in unnecessary search by generating candidates that Graphplan can prune, using memos, in layers that are not constructed explicitly by Stan. In fact, Stan generates no more candidates than Graphplan generates goal sets at layer $FP+1$. Indeed, Stan achieves a dramatic reduction in search by exploiting the correspondence between the goal trees generated at layers $n$ and $n-1$, demonstrated by Lemma 1. Because of this correspondence there is no need to construct the layers between $FP + 1$ and $n$ explicitly, and undertake all of the concommitant search from those layers.

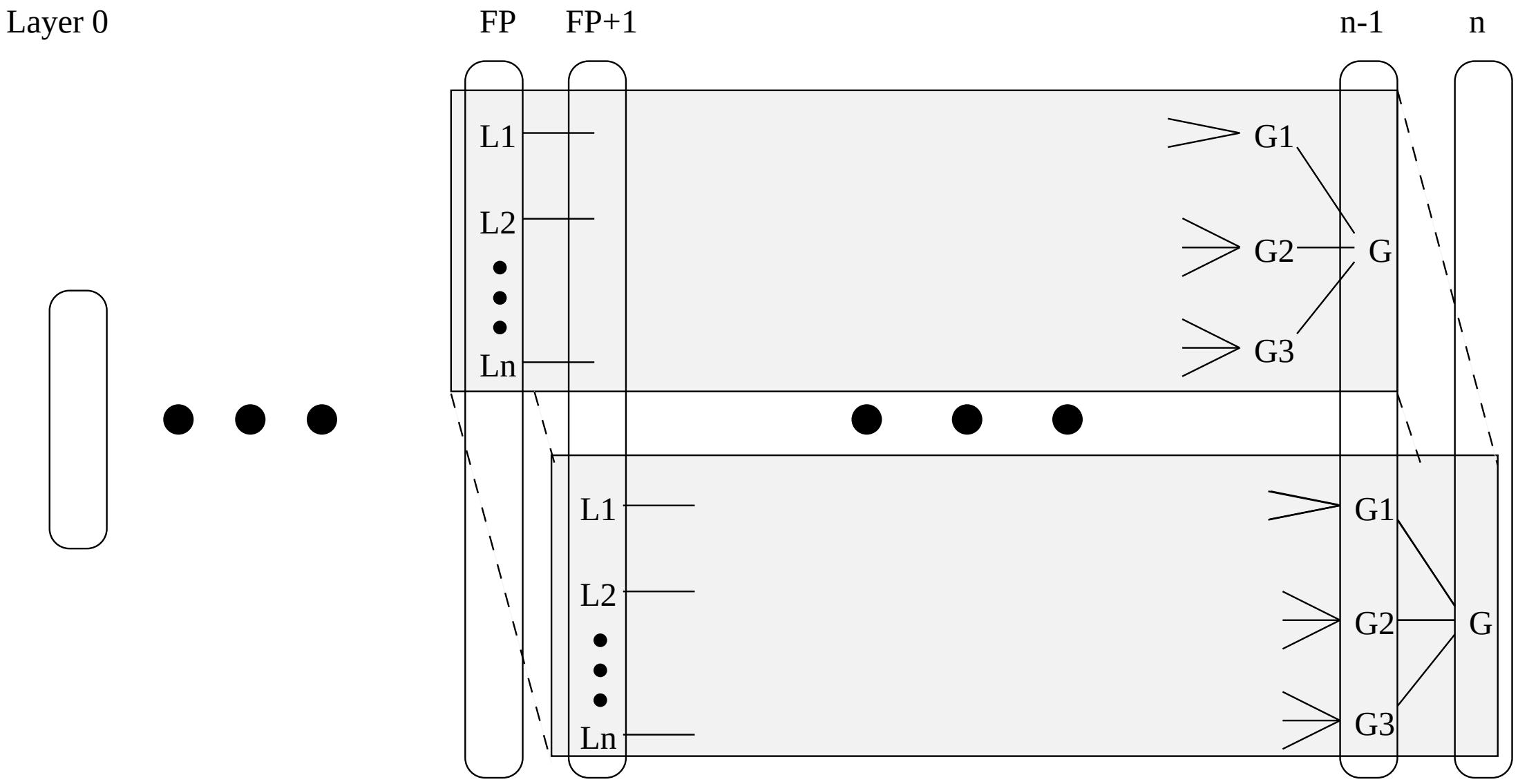

Figure 11: The sliding window of layers between $FP + 1$ and $n$.

Graphplan rebuilds the *sliding window*, shown in Figure 11, of layers between $FP$ and $n - 1$ as layers $FP + 1$ through to $n$. Stan simply promotes the leaves of the tree, generated at layer $FP$ in $GT_{n-FP,G,n-1}$, into layer $FP + 1$.

It is straightforward to show that the wave front maintains soundness. The search that Graphplan performs generates a goal tree of goal sets, as defined in Definition 12. In the example in Figure 10, the tree is rooted at G, with G1, G2 and G3 its children and G4 and G5 the children of G1. It can be seen from the picture that the tree structure generated by Graphplan, in which each successive layer would be embedded in a separate layer of the explicitly constructed graph, appears in a spiral of related goal sets between the fix point and buffer layers. All of the candidate goal sets lie in this same search tree and therefore no additional goal sets are generated. Graphplan constructs the final plan by reading off the sequence of action choices at each layer in the final graph. In Stan, the plan is obtained by reading off the initial fragment of the plan in the same way, from the layers preceding the fix point. The rest of the plan is extracted from the spiral. This extraction process yields the same path of action choices from the top level goal set to the candidate goal set as would be recorded explicitly in the Graphplan plan graph.

The only question remaining to be considered is whether the wave front has the same termination properties as Graphplan. It can be seen that it does since, if no new unsolvable goal sets are generated at the fix point, the queue will become empty and the planner terminates. This corresponds exactly to the termination conditions of Graphplan.

A subtlety concerns the interaction between the wave front and the subset memoization discussed in Section 2.2. In principle, subset memoization could cause the loss of all three of the desired properties of the graph. The way that Stan generates candidate goal sets is by simultaneously generating a candidate set whenever a goal set is memoized at the fix point. If the candidate set and the memoized set are one and the same, then the memoization of a *subset* of a goal set will lead to the propagation of only a subset of the actual candidate goals into the buffer and soundness might be undermined. If we use subset memoization at the wave front then the question arises whether sets that *contain* a memoized subset should be propagated forward as candidates. If they are not, then completeness is potentially lost, since there might be action sequences that could have been constructed following propagation that will not now be found. If they are, then termination is potentially lost, since the set that led to the construction of the memoized subset might itself be generated as a candidate. This could happen, for example in Figure 10, if G1 is unsolvable at the fix point but is generated again by consideration of a later candidate at the buffer.

To avoid these problems we have restored full subset memoization at the wave front. An alternative solution, which we are currently exploring, is to separate the subsets of goals memoized from the identification of the candidate sets. Both solutions avoid the loss of soundness because candidates are constructed from entire goal sets rather than from subsets. In the first solution, termination is preserved because memoizing full goal sets ensures that repeated candidates can be correctly identified as they recur. In the second solution, we would separately memoize candidates as they were generated to avoid repeated generation, thereby maintaining termination. In both cases, completeness is preserved by propagating goal-sets forward as new candidates provided only that they do not contain previously encountered candidates as subsets. If a potential candidate is a superset of an entire memoized candidate then it is correct not to propagate that potential candidate into the buffer because if the memoized candidate cannot be solved at the buffer then no superset of it can be solved there either.

## 5. Experimental Results

The results presented here use Stan version 2 (available at our website). We have performed experiments comparing Stan with and without the wavefront in order to demonstrate the advantages obtained by the use of the wave front. We have performed further experiments to compare Stan with the competition version of Ipp. There are some minor discrepancies in the timing mechanisms of these two systems. Stan measures elapsed time for the entire execution, whereas Ipp measures user+system time for graph construction and search but not for parsing of the problem domain and instance. On a single user machine as used for these experiments the discrepancy is negligible.

The problem domains used in this section have been selected to emphasise the benefits offered by the wave front. The important characteristic is that there should be an early fix point relative to the length of plan as instances grow. In the comparisons with Ipp the wave front accounts for the trends in performance, although Stan employs a range of other mechanisms which give it some minor advantages. Amongst these is the Tim machinery, which we have not decoupled as the problem domains used are the standard typed ones so that no significant advantage is obtained from inferring type structures automatically. Only the resource invariants inferred by Tim are exploited by Stan version 2, and we have indicated where this gives us an advantage over Ipp. Our ablation data sets confirm that the wave front is the most significant component in the performance of Stan in these experiments.

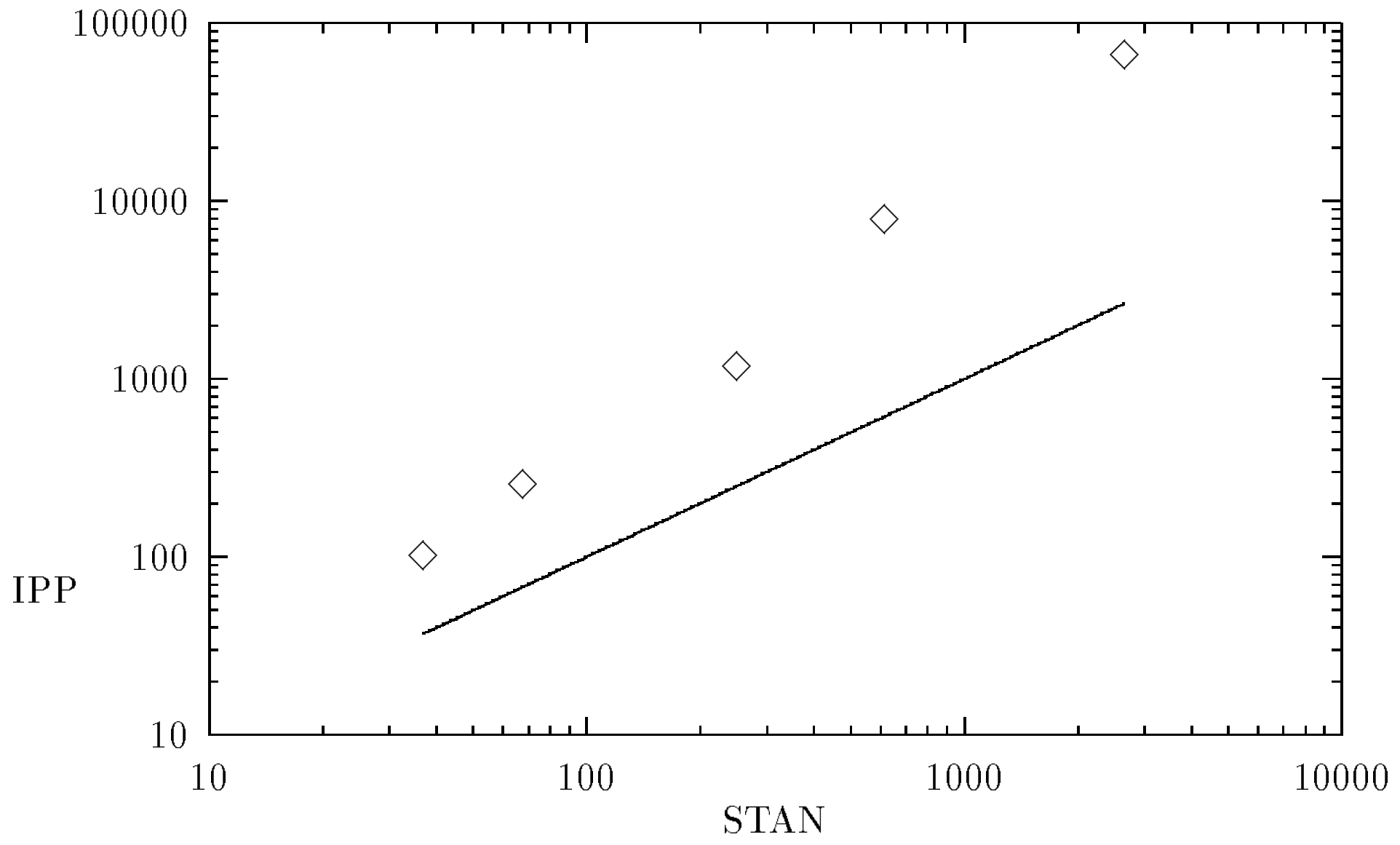

Figure 12: Stan compared with Ipp: solving Towers of Hanoi problems of 3-7 discs.

Stan is capable of efficiently solving larger Towers of Hanoi instances than are presented in the graph in Figure 12, which accounts for the additional point in Figure 13. Stan with the wave front found the 511-step plan for the 9-disc problem in less than 7 minutes using about 15Mb of memory. During the experiments reported here, Ipp was terminated after 15 minutes having reached layer 179 out of 255 layers in the 8-disc problem. We observe that on a machine with 1Gb of RAM, Ipp has solved this problem in 8 minutes.

The results for the Gripper domain demonstrate only a small advantage for Stan. The reason is because the search space grows exponentially in the size of the graph in the Gripper domain, so that the cost of searching dominates everything else. Although the search spaces for Towers of Hanoi instances also grow exponentially, they grow as $2^x$ whereas Gripper instances grow as $x^x$ (where $x$ is the number of discs or balls respectively). Although the wave front helps under these conditions, the size of the search space dwarfs the benefits it offers. The Ferry domain is a less rapidly growing version of the gripper domain since only one vehicle can be carried on each journey, reducing the number of choices at each layer. The difference in benefits obtained in the Towers of Hanoi domain relative to the Gripper and Ferry domains can be explained by consideration of the table in Figure 16. The benefits of the wave front are proportional to the number of layers which exist implicitly between the buffer and the layer from which the plan is ultimately found. In the Towers of Hanoi the number of implicit layers is exponential in the number of discs whereas the number of layers between the initial layer and the buffer is linear in the number of discs. Therefore the benefits offered by the use of the wave front are magnified exponentially as the problem instance grows. On the other hand, in both Gripper and Ferry there is only a linear growth in the difference between plan length and fix point layer, so benefits are magnified only linearly. This analysis can be confirmed by observation of Figures 12, 14 and 17.

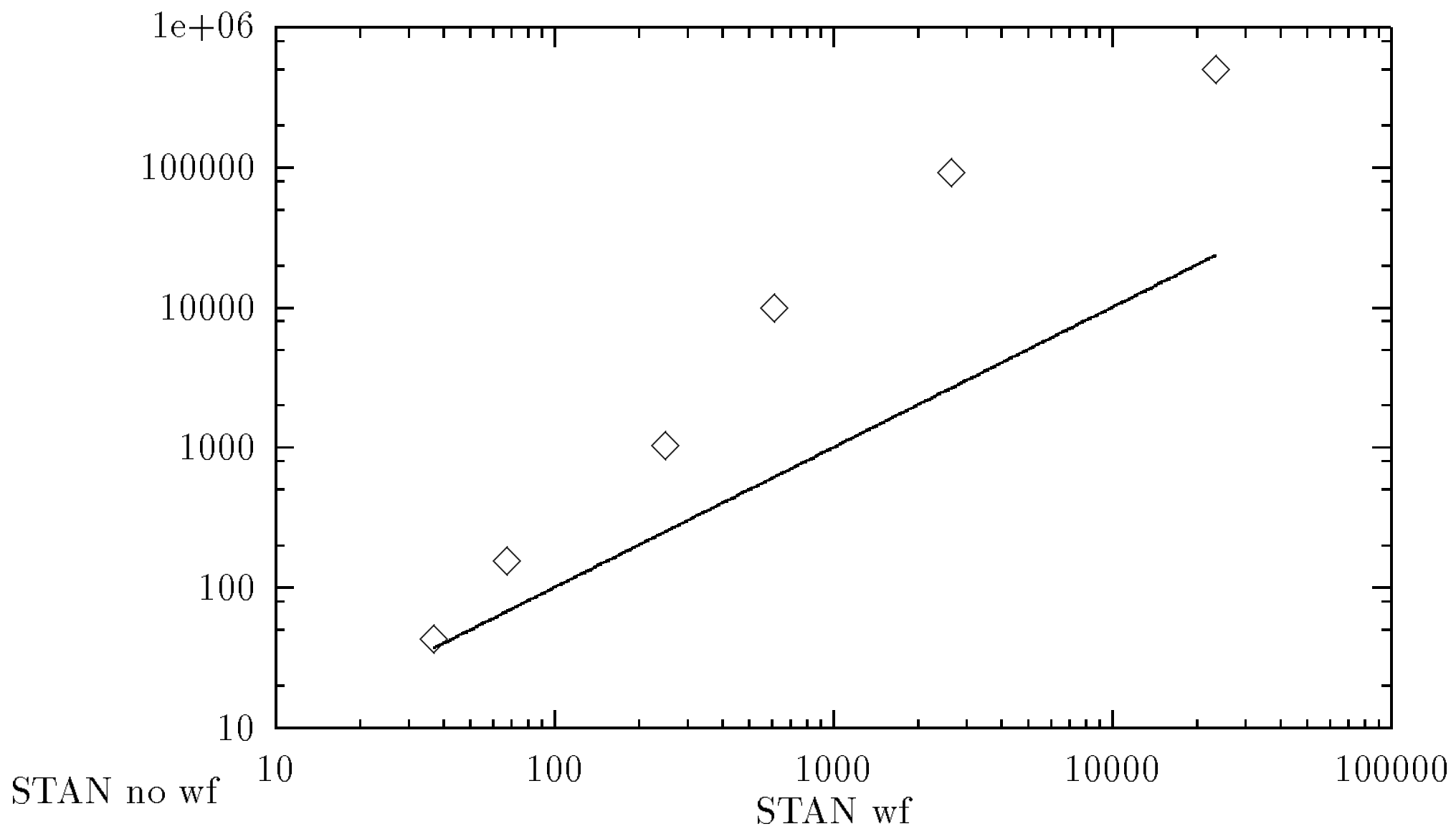

Figure 13: Stan with and without the wave front: solving Towers of Hanoi problems of 3-8 discs.

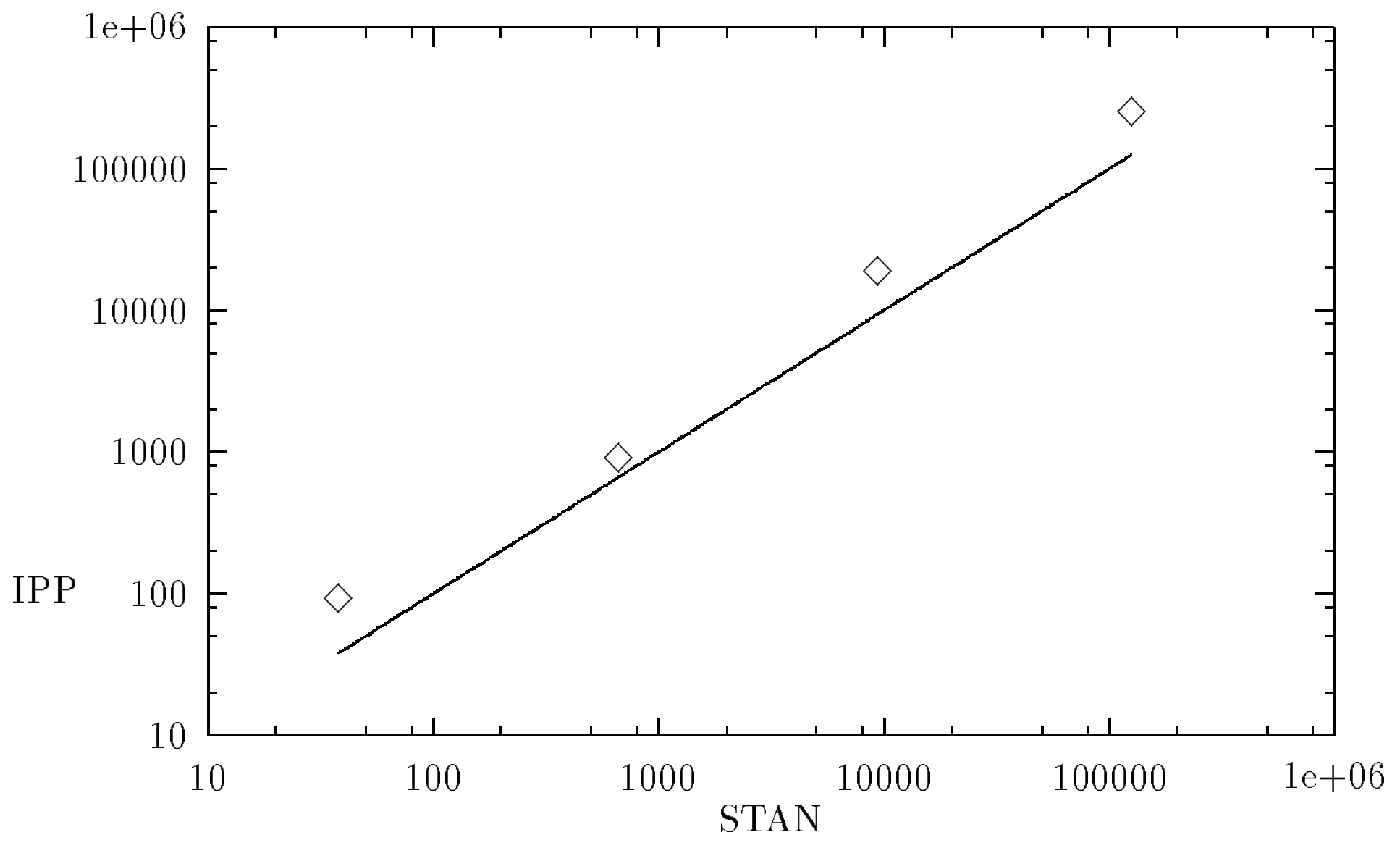

Figure 14: Stan compared with Ipp: solving Gripper problems of 4-10 balls.

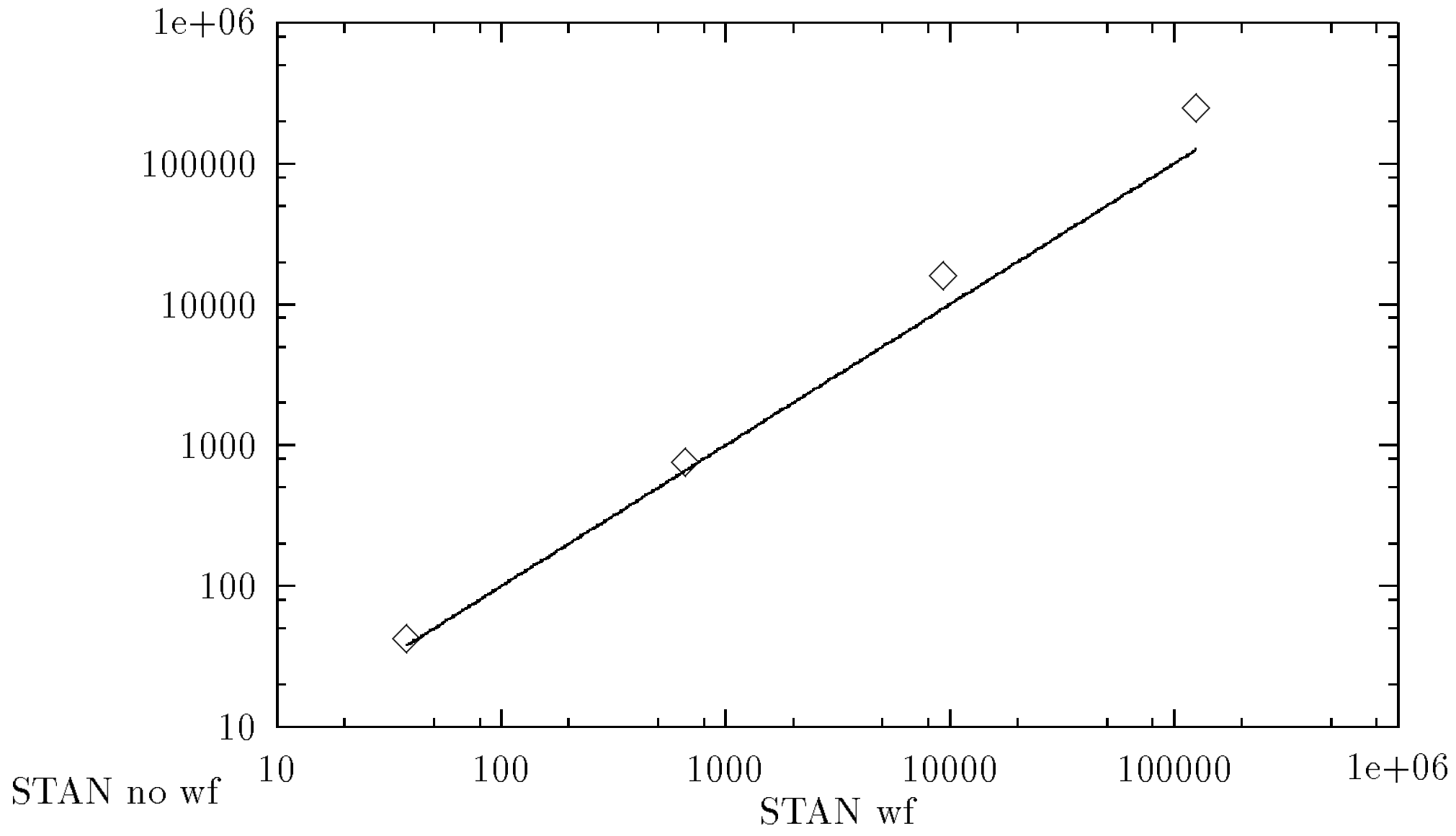

Figure 15: Stan with and without the wave front: solving Gripper problems of 4-10 balls.

| Domain | Parameter n | Plan Length | Buffer |
|---|---|---|---|
| Towers of Hanoi | no. discs | $2^n - 1$ | $n + 3$ |
| Gripper | no. balls | $2n - 1$ | 5 |
| Ferry | no. vehicles | $4n - 1$ | 7 |
| Complete-Graph TSP | no. cities | $n$ | 4 |

Figure 16: Relative values of plan length and number of layers to buffer for four domains.

The benefit of the wave front is measured not only in terms of the cost of construction that is avoided by not explicitly building the layers beyond the buffer, but also in terms of the search that is avoided in those layers. Crudely, the benefits can be measured as the number of layers not constructed multiplied by the search effort avoided at each of those layers. Thus, the number of layers not constructed magnifies the benefits obtained by not searching amongst them. This is a simplification, since the search effort avoided at successive layers increases as they get further away from the fix point, but it gives a guide to the kind of benefits that can be expected from the wave front.

Stan obtains significant advantages over Ipp in the Complete-Graph Travelling Salesman domain, as Figure 19 demonstrates. Some of these advantages are obtained by ex-ploiting the resource analysis techniques of Tim (Fox & Long, 1998), whilst a significant proportion of the advantage is obtained from the use of the wave front, as Figure 20 shows. Resource analysis allows a lower bound to be determined on the number of layers that must be built in a plan graph before it is worth searching for a plan. In the Complete-Graph Travelling Salesman domain this is very powerful, as the calculated bound is $n$, the number of cities in the instance, which is precisely the correct plan length. In this domain, if no search is done until $n$ layers are constructed, no search needs to be done at all since it doesn't matter in what order the cities are visited. This would allow the problem to be solved in polynomial time (of course, this only makes sense because the Complete-Graph TSP used here is simpler than the NP-hard TSP). However, when the wave front is used, the buffer is at layer 4 and the only way of finding the plan is to generate all of the candidate goal sets at layer 4, of which there are an exponential number. The use of the wave front in this domain therefore forces Stan to take exponential time in the size of the instances. Despite this the wave front offers great advantages. The benefits increase exponentially as instance sizes grow although the magnification of these benefits at each layer is only linear, see Figure 16, although the benefits are offset by the exponential growth in the number of candidates. It must be observed that in Figure 19, the figures are extrapolated for Ipp for instances in which $n$ is greater than 14. The extrapolation was based on Ipp's performance on instance sizes between 2 and 14, which demonstrates a clear exponential growth.

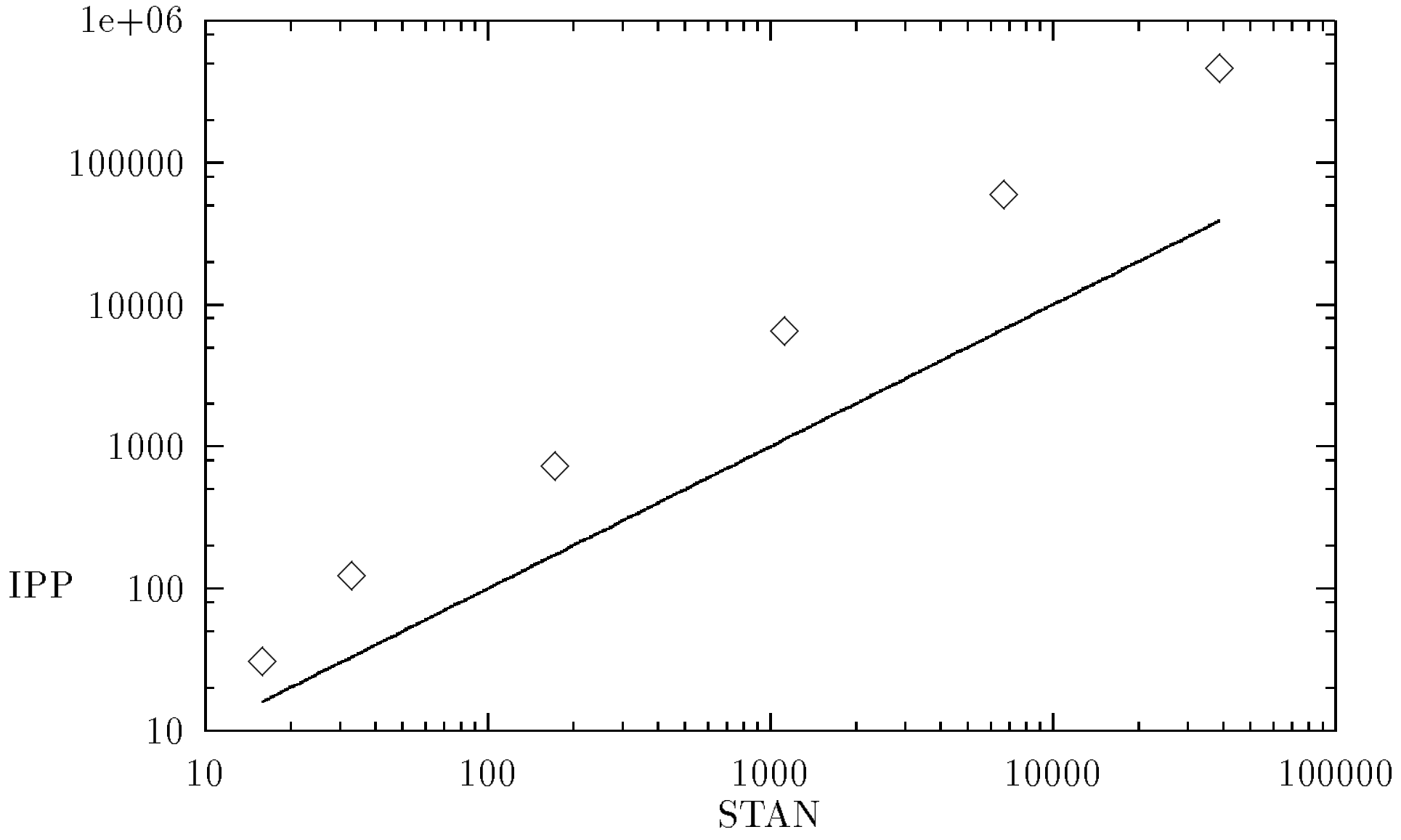

Figure 17: Stan compared with Ipp: solving Ferry problems of 2-12 cars.

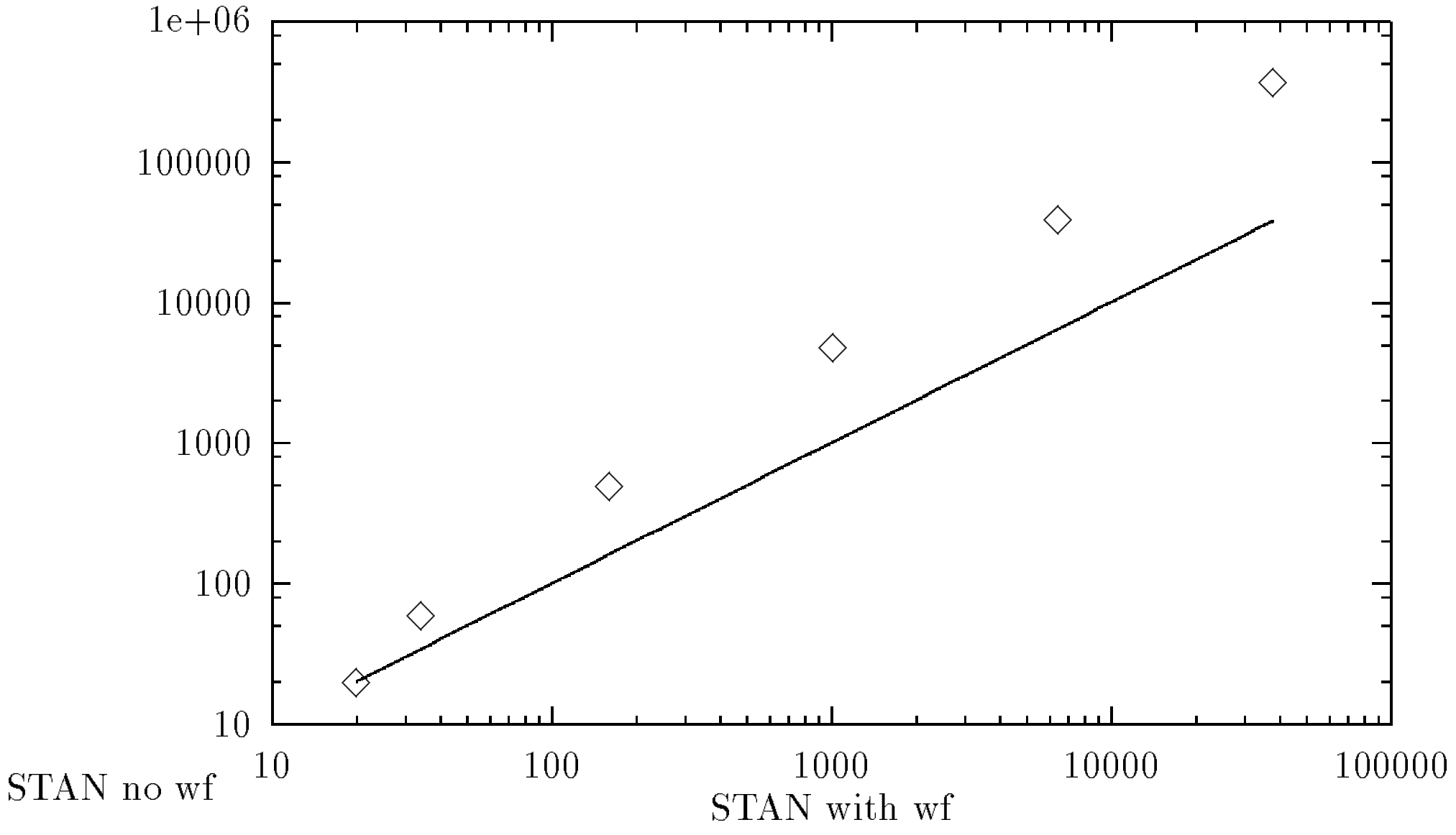

Figure 18: Stan with and without the wave front: solving Ferry problems of 2-12 cars.

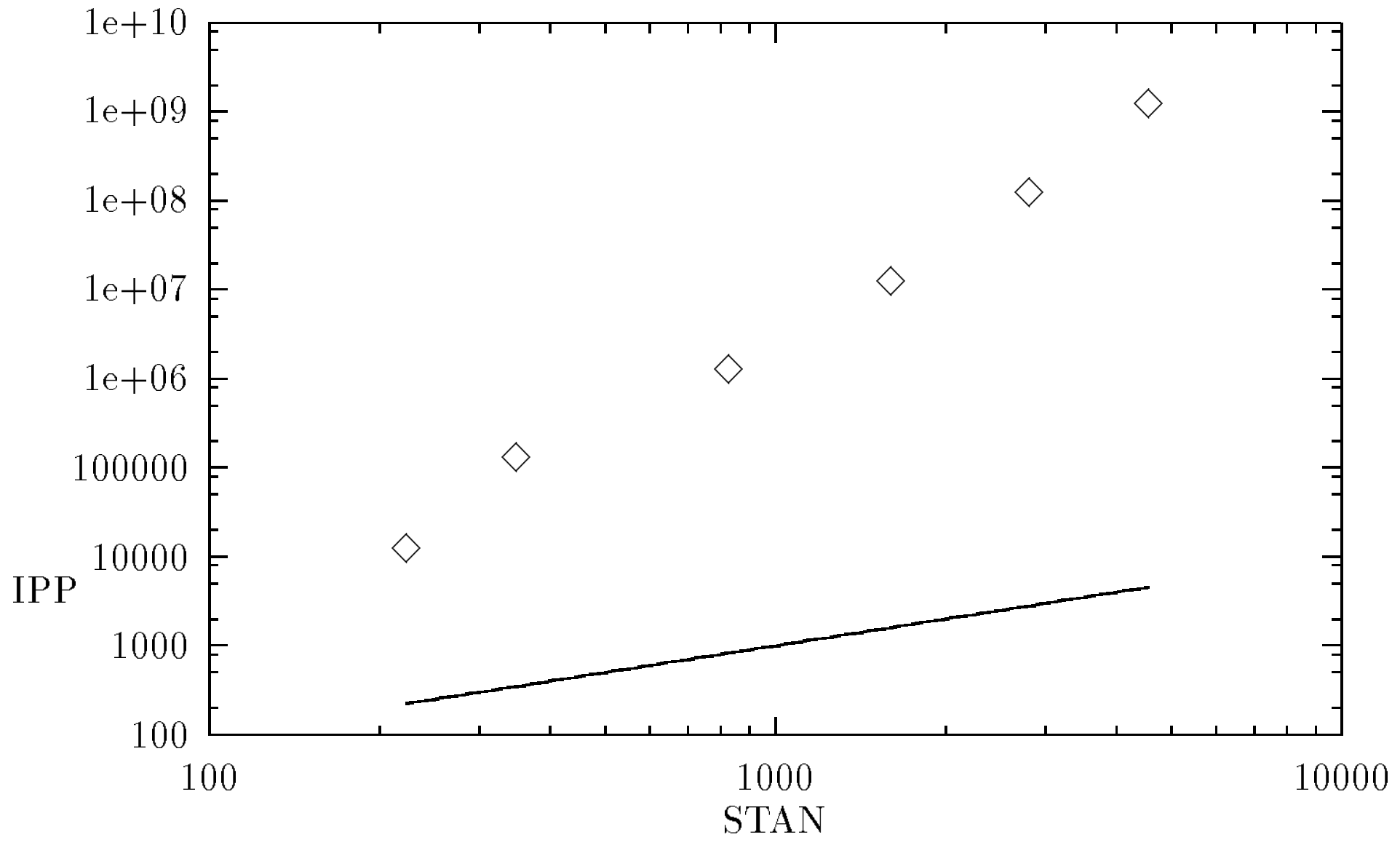

Figure 19: Stan compared with Ipp: solving Complete-Graph Travelling Salesman problems of 10-20 cities.

It appears that we could allow the resource analysis to over-ride the wave front when a domain is encountered in which it can be guaranteed that explicit construction of the graph will be more efficient. In practice the Complete-Graph Travelling Salesman domain seems exceptional, since search is eliminated if the graph is constructed to layer $n$, and if this were not the case the explicit construction and subsequent search would be more costly than the use of the wave front.

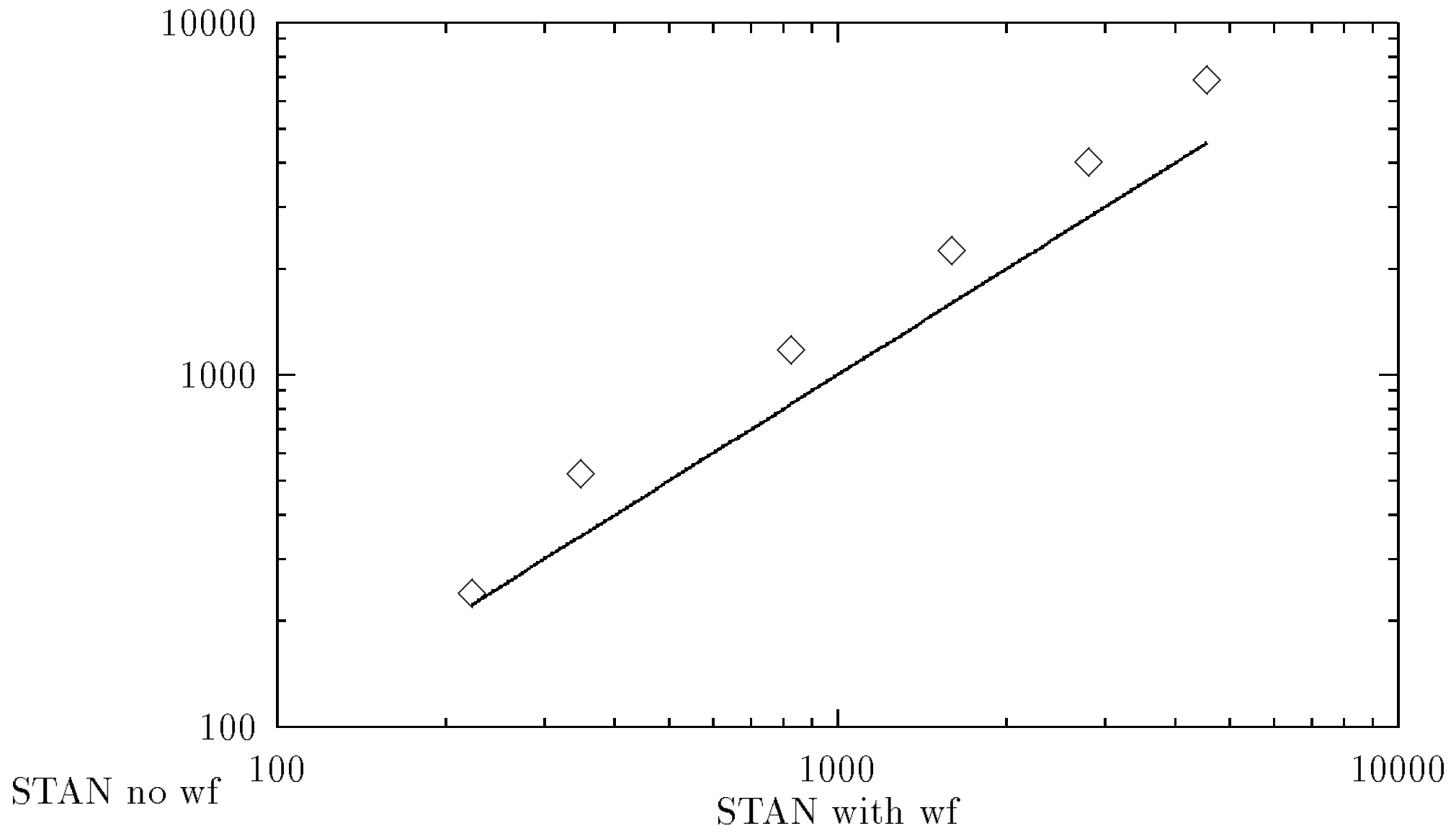

Figure 20: Stan with and without the wave front: solving Complete-Graph TSP problems.

### 5.1 The Wave Front Heuristic

The queue of candidate goal sets considered in the buffer can be implemented as an unordered structure in which goal sets are selected for consideration according to more sophisticated criteria than the order in which they were stored. In principle, this could save much searching effort since it could avoid costly consideration of goal sets which turn out to be unsolvable before meeting a solvable goal set. We have experimented with a number of goal set selection heuristics which favour goal sets for which the search progresses deepest into the graph structure. These sets are considered to be closer to being solvable than sets which fail in a layer very close to the buffer. Candidates are evaluated by considering the length of the plan fragment associated with the candidate and the extent to which the failed search penetrated into the graph when initiated from the fix point layer when the candidate was first generated. The search penetration should be maximized while the plan fragment length should be minimized. Considering the goal sets in some order other than that in which they are generated does not affect any of the formal properties of the planner other than the optimality of the plans generated. Non-optimal plans can be favoured because the balance between fragment length and penetration can cause candidates with shorter fragments to be overlooked.

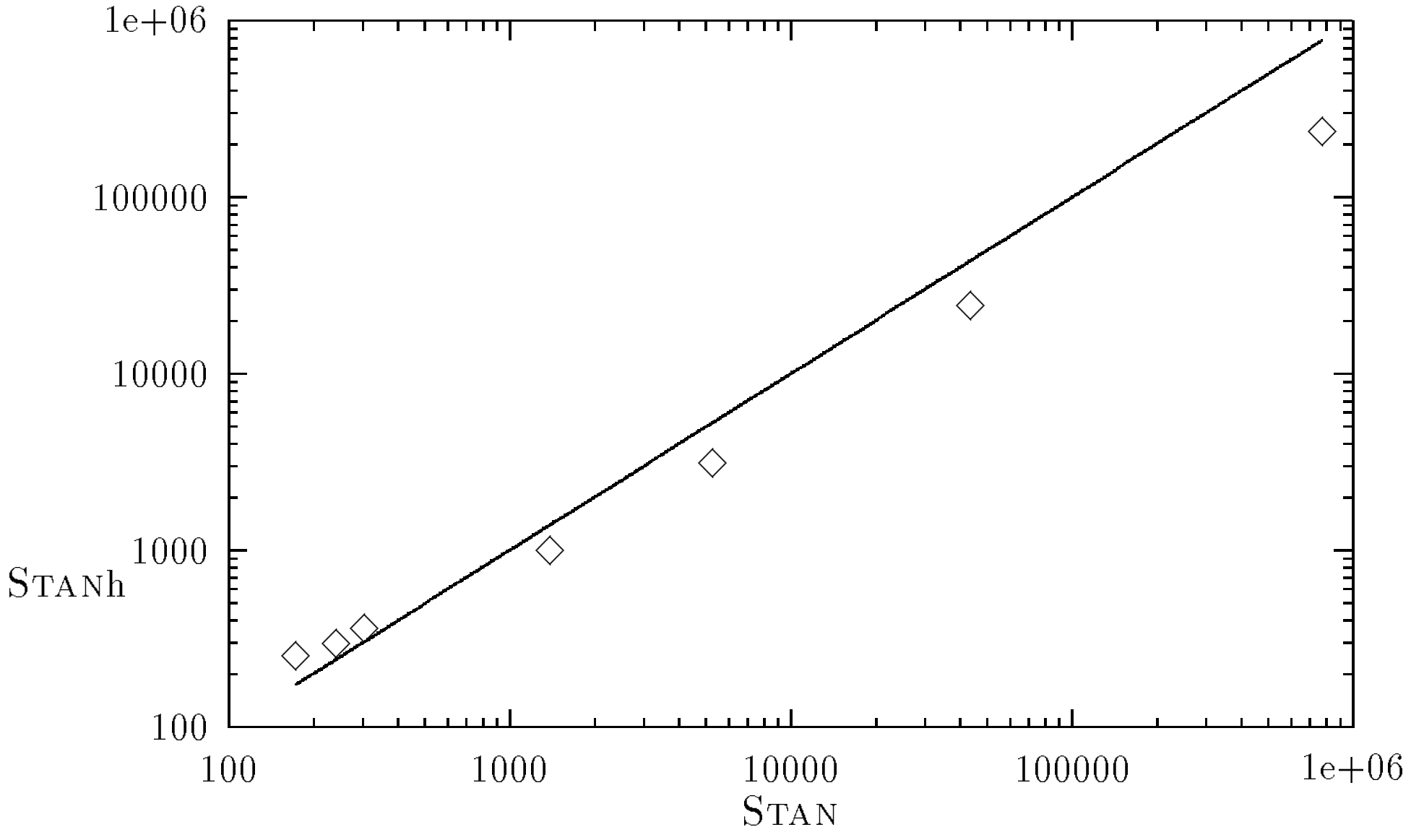

Figure 21: Towers of Hanoi with (StanH) and without the heuristic: 3-9 discs.

Using the heuristic Stan is able to solve Towers of Hanoi problems very efficiently, as Figure 21 shows. As previously, the graph is log-log scaled. The line indicates at least a polynomial improvement in the size of instances. The heuristic was originally developed by consideration of blocks world problems, in which it also performs well. However, it does not provide a reliable advantage so it is not used in Stan version 2. It *was* used on all problems in the competition but often represented a heavy overhead for Stan. We are continuing to experiment with alternative domain-independent evaluation criteria.

## 6. Conclusion

This paper presents two improvements on the representation of the plan graph exploited by Graphplan-based planners. These are: the representation of the graph as a single pair of layers, called a *spike*, built around bit vectors and logical operations, and the use of a *wave front* which avoids the explicit construction of the graph beyond the fix point. We describe a highly efficient procedure for checking mutex relations between actions and explain what characteristics of problems allow its full exploitation. The spike and the wave front have both been implemented in Stan, a Graphplan based planner version 1[1] of which competed successfully in the AIPS-98 planning competition. We have presented empirical evidence to support both improvements. The first set of data demonstrates the increase in graph

1. Version 1 contained implementations of both the spike and the wave front. Version 2 enhances both of these mechanisms with improved implementation and the addition of the *changedActs* mechanism discussion in Section 2.1.

construction efficiency obtained by the use of the spike. The second set of data shows the advantages obtained during the search of the plan graph by using the wave front.

Stan also employs the state invariant inference machinery of Tim (Fox & Long, 1998), but in version 2 the integration of the invariants into the graph construction process is still only partial. We observe that the mutex relations generated in the Complete-Graph TSP, in particular, are almost entirely domain invariants of the kind inferred by Tim. Integration of these inferred invariants into the graph would allow these mutex relations to be identified immediately as permanent and eliminate them from retesting, dramatically enhancing Stan's graph construction performance in this domain. A similar advantage would be obtained across other domains since many of the mutex relations inferred during graph construction correspond to invariants of the various forms inferred efficiently by Tim during a preprocessing stage.

## Appendix A. Website Addresses

Online Appendix 1 contains a complete collection of the domains and problems used in this paper, executables (Linux and Sparc-Solaris binaries) for Stan and the reduced version of Stan for graph construction, and a *diff* file showing how the graph constructing version of IPP was generated.

The results of the AIPS-98 planning competition can be found at:
`http://ftp.cs.yale.edu/pub/mcdermott/aipscomp-results.html`.

The Stan website can be found at:
`http://www.dur.ac.uk/~dcs0www/research/stanstuff/planpage.html`.